\pdfoutput=1
\documentclass[letterpaper]{article}

\usepackage{aaai2027}
\usepackage{amsmath}
\usepackage{amssymb}
\usepackage{bm}
\usepackage{booktabs}
\usepackage{graphicx}
\usepackage{float}
\usepackage{url}
\usepackage{caption}

\nocopyright

\title{FBFM: A Training-Free Asynchronous Feedback Mechanism for Flow-Matching in
World-Action Models Execution}
\author{
Peize Li\textsuperscript{\rm *,1} \quad
Ruimeng Zhang\textsuperscript{\rm *,2,3,4,5} \quad
Ru Zhang\textsuperscript{\rm 5,6} \quad
Cong Huang\textsuperscript{\rm 1,2,3}\\
Kai Chen\textsuperscript{\rm \textdagger,1,2,3} \quad
Shanghang Zhang\textsuperscript{\rm \textdagger,4}
}
\affiliations{
\textsuperscript{\rm 1}DeepCybo \quad
\textsuperscript{\rm 2}Zhongguancun Academy \quad
\textsuperscript{\rm 3}Zhongguancun Institute of Artificial Intelligence\\
\textsuperscript{\rm 4}Peking University \quad
\textsuperscript{\rm 5}Beijing Institute of Technology \quad
\textsuperscript{\rm 6}Tsinghua University\\
\textsuperscript{\rm *}Equal contribution \quad
\textsuperscript{\rm \textdagger}Corresponding authors
}
\date{}

\begin{document}

\maketitle


\begin{abstract}
Although world-action models (WAMs) enhance long-horizon robot control by
predicting visual evolution before acting, long-horizon reliability demands
repeated re-grounding in real observations---not recursive rollout. Existing WAMs
address this by refreshing history or KV cache with ground-truth data between
chunks. However, such chunk-wise feedback operates at a coarse temporal
granularity and thus fails to correct prediction errors at the individual
time-step level. To address this, we propose Feedback Flow Matching (FBFM), a
training-free inference mechanism that pushes re-grounding inside the actively
generated chunk. During flow matching, FBFM applies a masked pseudoinverse
correction to the conditional velocity field: it leverages the preceding action
chunk to guide generation of the next action chunk, and uses the image observed
after executing that preceding chunk to guide the next frame prediction. This
cross-chunk pairing---where feedback from one chunk arrives in time to shape the
next---creates an asynchronous loop that corrects errors without waiting for
chunk boundaries. Being training-free, the mechanism improves responsiveness to
unexpected events and suppresses drift in long-horizon tasks. We evaluate FBFM
on a joint-generation WAM (DreamZero) and a stage-wise WAM (LingBot-VA). Across
42 selected RoboTwin2.0 tasks, FBFM raises the equal-weight task-configuration
success rate by 3.0 percentage points; across four LIBERO suites, it provides a
modest 0.6-point pooled gain, including 2.5-point improvements on LIBERO-Goal
and LIBERO-10. Prediction from recorded physical-robot observations further
shows improved state tracking. We argue that FBFM offers a new paradigm for
fine-grained online correction, bridging open-loop flow generation with
closed-loop real-world dynamics.
\end{abstract}

\begingroup
\renewcommand{\thefootnote}{}
\footnotetext{\hspace{-1.8em}Ruimeng Zhang contributed as an intern at Zhongguancun Academy and Zhongguancun Institute of Artificial Intelligence, and as a Peking University RA; Ru Zhang contributed as a Tsinghua University RA.}
\endgroup


\section{Introduction}
\label{sec:introduction}


World--action models (WAMs) have emerged as a promising paradigm for long-horizon
robotic manipulation. Rather than producing actions solely from the current
observation and task instruction, a WAM explicitly models how the scene may evolve
and uses the predicted future to support action generation. This look-ahead
structure is particularly attractive for tasks involving multi-step reasoning,
contact-rich interaction, or object rearrangement. Yet, the predicted future that
enables such reasoning also creates a deployment risk: once state prediction
deviates from the actual physical evolution, subsequent actions may be conditioned
on states that the robot will never reach. Reliable WAM execution therefore
requires not only longer-horizon prediction, but repeated re-grounding in
observations revealed during execution.

Existing WAM systems partially address this requirement by refreshing their
history or internal context with ground-truth observations between action chunks.
Such context updates improve the condition available to a subsequent generation
call, but the returned observations remain conditioning information rather than
measurements imposed on selected variables of the future currently being
generated. This distinction creates a temporal-granularity gap: chunk-level
re-grounding cannot correct state-prediction errors at individual time steps of an
active chunk. As a result, the future used to produce actions may remain
inconsistent with physical transitions that have already occurred.

Work on asynchronous action execution provides a useful starting point. Real-Time
Chunking (RTC) shows that actions inherited from an executing chunk can constrain
temporally aligned coordinates of the next action chunk through training-free
inference-time inpainting~\cite{black2025realtime}. This construction preserves
action continuity under inference delay, but its constrained variables lie in
action space. It does not use newly observed state transitions to re-ground the
explicit future-state trajectory generated by a WAM. Acting consistently across a
chunk boundary and maintaining a future that is consistent with the environment
are therefore complementary requirements.

The WAM-specific opportunity arises from the overlap between execution and
generation. While the preceding action chunk is being executed, a new state--action
chunk may still be undergoing Flow Matching. Executing
$a_{t+i-1}^{\mathrm{prev}}$ induces a real environment transition, whose
observation can be encoded as $z_{t+i}$. This latent state corresponds to a
particular temporal slot of the generated future-state chunk rather than to an
undifferentiated summary of past error. The execution window therefore turns WAM
inference into a dynamically partially observed generation problem: state slots
revealed by the environment should be anchored to their measured values, while
unobserved slots should remain governed by the pretrained WAM prior.

We propose Feedback Flow Matching (FBFM), a training-free inference mechanism that
expresses both state feedback and cross-chunk action consistency as time-aligned
constraints on Flow Matching. At each solver evaluation, newly available latent
states are exposed to the active chunk through a dynamically updated state target
and mask. Complementarily, actions committed by the preceding chunk define a fixed
target over the aligned action overlap. A masked pseudoinverse discrepancy and its
vector--Jacobian product correct the corresponding sampling velocity without
updating the WAM parameters. The two constraints share one mathematical interface
but retain different semantics: state feedback incorporates newly measured
physical transitions, whereas previous-action consistency preserves commitments
already made by the controller.

FBFM applies to both stage-wise generation and joint generation. In a stage-wise
WAM, feedback first guides the state flow, and the latest corrected state context
conditions an action flow whose cross-chunk overlap is constrained by the preceding
actions. State feedback arriving after state generation can still refresh the
context used by the ongoing action flow. In a joint-generation WAM, state and
action targets constrain a shared flow; through the cross-modal blocks of the
clean-endpoint Jacobian, a state residual can directly correct action coordinates
during the same solver evaluation. FBFM thus introduces fine-grained closed-loop
correction while preserving the pretrained generation process of either WAM
architecture.

This paper makes three contributions:
\begin{enumerate}
    \item We formulate chunk-internal WAM feedback as a dynamically partially
    observed generation problem that preserves both the values and temporal
    locations of returned state observations.
    \item We introduce FBFM, a training-free masked pseudoinverse-guidance mechanism
    that combines dynamic latent-state feedback with fixed cross-chunk action
    consistency while keeping the pretrained WAM frozen.
    \item We instantiate FBFM in both stage-wise and joint-generation WAMs,
    evaluate it on selected LIBERO and RoboTwin2.0 tasks, and study state tracking
    on observation sequences recorded from a physical robot task.
\end{enumerate}


\section{Preliminaries and Notations}
\label{sec:preliminaries}


\subsection{Problem Formulation: Agent--Environment Interaction}
\label{sec:preliminaries:problem}

We consider an embodied agent interacting with an environment over a family of tasks
$\mathcal{T}$. The interaction is modeled as a controlled Markov process
\begin{equation}
    \mathcal{M} = (\mathcal{S}, \mathcal{A}, P),
    \label{eq:controlled-markov-process}
\end{equation}
where $\mathcal{S}$ and $\mathcal{A}$ denote the environment state space and the
action space, respectively, and $P(s_{t+1}\mid s_t,a_t)$ is the environment
transition kernel. Each task $T\in\mathcal{T}$ may specify an initial-state
distribution $\rho_T$ and a task condition $c_T$, such as a language instruction
or a goal specification. We assume that the tasks share the same state space, action
space, and physical dynamics. At environment time step $t$, the environment is in
state $s_t\in\mathcal{S}$, the agent applies an action $a_t\in\mathcal{A}$, and
the next state evolves according to
\begin{equation}
    s_0\sim\rho_T,
    \qquad
    s_{t+1}\sim P(\cdot\mid s_t,a_t).
    \label{eq:environment-transition}
\end{equation}

The physical state $s_t$ is not necessarily exposed directly to the agent. Let
$\mathcal{O}$ denote the sensing process and $E$ a perceptual encoder. We define
the latent state available to the world--action model (WAM) as
\begin{equation}
    z_t = E\!\left(\mathcal{O}(s_t)\right),
    \label{eq:latent-state-encoding}
\end{equation}
where $z_t\in\mathcal{Z}$ summarizes the sensory observation of $s_t$. This
distinction allows the environment dynamics to remain Markovian in $s_t$, while
the agent may act from encoded and potentially partial observations. We denote the
interaction history available at time $t$ by
\begin{equation}
    \mathcal{H}_t = (c_T,z_0,a_0,z_1,a_1,\ldots,a_{t-1},z_t),
    \label{eq:interaction-history}
\end{equation}
and write the action-selection process generally as
$a_t\sim\pi(\cdot\mid\mathcal{H}_t)$. This notation does not require $z_t$ itself
to be a sufficient Markov state.

Rather than selecting only a single action, a WAM predicts future latent states and
actions over a horizon $H$. Starting at time $t$, we denote its predicted latent
and action chunks by
\begin{equation}
    \begin{aligned}
    \hat{\mathbf{Z}}_{t,H}
    &= (\hat z_{t+1},\ldots,\hat z_{t+H}),\\
    \hat{\mathbf{A}}_{t,H}
    &= (\hat a_t,\ldots,\hat a_{t+H-1}),
    \end{aligned}
    \label{eq:wam-prediction-chunks}
\end{equation}
with the generic predictive model
\begin{equation}
    (\hat{\mathbf{Z}}_{t,H},\hat{\mathbf{A}}_{t,H})
    \sim p_\theta(\cdot\mid\mathcal{H}_t).
    \label{eq:wam-predictive-model}
\end{equation}

This expression does not prescribe how the two chunks are generated: a WAM may
model them with a joint generative process or factorize generation into successive
latent-state and action stages. During execution, realized transitions and external
disturbances can cause the newly encoded latent $z_{t+1}$ to deviate from its
prediction $\hat z_{t+1}$. Our objective is therefore not to learn a
reward-maximizing policy, but to incorporate such newly available state feedback,
together with previously generated action constraints, into the inference process of
a pretrained WAM. Accordingly, rewards, returns, and value functions are not part of
our formulation.

\subsection{Flow Matching}
\label{sec:preliminaries:flow-matching}

We distinguish an individual variable at one environment step from a chunk spanning
multiple steps. An action
$a_t\in\mathbb{R}^{d_a}$ is a single control command at environment time $t$,
whereas
\begingroup
\small
\begin{equation}
    \mathbf{A}_t
    = [a_t^{\mathsf T},a_{t+1}^{\mathsf T},\ldots,
       a_{t+H-1}^{\mathsf T}]^{\mathsf T}
    \in\mathbb{R}^{D_A},
    \qquad D_A=H d_a,
    \label{eq:action-chunk}
\end{equation}
\endgroup
is the time-stacked action vector of horizon $H$. Similarly,
$z_t\in\mathcal{Z}\subseteq\mathbb R^{d_z}$ denotes the single-step latent state
defined above, while
\begingroup
\small
\begin{equation}
    \mathbf{Z}_t
    = [z_{t+1}^{\mathsf T},z_{t+2}^{\mathsf T},\ldots,
       z_{t+H}^{\mathsf T}]^{\mathsf T}
    \in\mathbb R^{D_Z},
    \qquad D_Z=H d_z,
    \label{eq:latent-chunk}
\end{equation}
\endgroup
denotes the time-stacked future latent-state vector aligned with the action horizon.
Bold uppercase symbols therefore denote chunk-level vectors throughout the paper.
We use
lowercase $x$ for a generic single-step variable and uppercase
$\mathbf{X}$ for its chunk-level counterpart. Thus,
$(x,\mathbf{X})=(a,\mathbf{A})$ for action generation and
$(x,\mathbf{X})=(z,\mathbf{Z})$ for latent-state generation. For a WAM that
generates both modalities jointly, $\mathbf{X}_t$ may instead denote their
concatenation,
\begin{equation}
    \mathbf{X}_t
    =
    \begin{bmatrix}
        \mathbf Z_t\\
        \mathbf A_t
    \end{bmatrix}
    \in\mathbb R^{D_X},
    \qquad D_X=D_Z+D_A.
    \label{eq:joint-wam-chunk}
\end{equation}
For bold vectors $\mathbf x\in\mathbb R^{d_x}$ and
$\mathbf y\in\mathbb R^{d_y}$, all derivatives use the numerator-layout
Jacobian convention,
$\partial\mathbf y/\partial\mathbf x\in\mathbb R^{d_y\times d_x}$.
When $\mathbf X$ denotes a generic generation vector, $D$ denotes its
corresponding dimension.

This generic notation does not imply a particular WAM factorization. A joint WAM
transports the concatenated $\mathbf{X}_t$ with one flow, whereas a stage-wise WAM
applies separate flows to $\mathbf{Z}_t$ and $\mathbf{A}_t$. In either case, the
elements of a generated chunk remain aligned with individual environment steps; this
alignment will later allow FBFM to constrain selected $z_{t+i}$ and
$a_{t+i}$, rather than treating the chunk as an indivisible unit.

Flow Matching learns a time-dependent vector field that transports samples from a
simple source distribution to the conditional data distribution. Let
$\tau\in[0,1]$ denote continuous flow time, distinct from the environment index
$t$. Given a target chunk $\mathbf{X}_t$ and Gaussian noise
$\boldsymbol{\epsilon}\sim\mathcal{N}(\mathbf{0},\mathbf{I})$ of the same dimension,
we use the linear conditional probability path
\begin{equation}
    \mathbf{X}_t^\tau
    = (1-\tau)\boldsymbol{\epsilon}+\tau\mathbf{X}_t,
    \qquad
    \mathbf{X}_t^0=\boldsymbol{\epsilon},
    \quad
    \mathbf{X}_t^1=\mathbf{X}_t.
    \label{eq:linear-flow-path}
\end{equation}
Its conditional target velocity is constant along the path:
\begin{equation}
    \mathbf{u}(\mathbf{X}_t^\tau\mid\mathbf{X}_t)
    = \frac{\mathrm{d}\mathbf{X}_t^\tau}{\mathrm{d}\tau}
    = \mathbf{X}_t-\boldsymbol{\epsilon}.
    \label{eq:conditional-flow-velocity}
\end{equation}

Conditioned on the interaction history $\mathcal{H}_t$, a Flow-Matching model
$v_\theta$ takes the current noisy chunk and flow time as inputs and predicts a
velocity with the same dimension as $\mathbf{X}_t$. It is trained using
\begin{equation}
    \mathcal{L}_{\mathrm{FM}}(\theta)
    =
    \mathbb{E}
    \left[
        \left\|
        v_\theta(\mathbf{X}_t^\tau,\tau;\mathcal{H}_t)
        -
        (\mathbf{X}_t-\boldsymbol{\epsilon})
        \right\|_2^2
    \right],
    \label{eq:flow-matching-objective}
\end{equation}
where the expectation is over
$(\mathcal H_t,\mathbf X_t)\sim\mathcal D$,
$\boldsymbol\epsilon\sim\mathcal N(\mathbf 0,\mathbf I)$, and
$\tau\sim p(\tau)$. Uniform sampling is the standard choice, while non-uniform
schedules may emphasize particular noise levels.

At inference time, generation starts from
$\mathbf{X}_t^0\sim\mathcal{N}(\mathbf{0},\mathbf{I})$ and solves the conditional
ordinary differential equation
\begingroup
\small
\begin{equation}
    \begin{aligned}
    \frac{\mathrm{d}\mathbf{X}_t^\tau}{\mathrm{d}\tau}
    &= v_\theta(\mathbf{X}_t^\tau,\tau;\mathcal{H}_t),\\
    \hat{\mathbf{X}}_t
    &= \mathbf{X}_t^0
    + \int_0^1
    v_\theta(\mathbf{X}_t^\tau,\tau;\mathcal{H}_t)\,
    \mathrm{d}\tau.
    \end{aligned}
    \label{eq:flow-matching-ode}
\end{equation}
\endgroup
For example, a forward Euler solver uses
\begin{equation}
    \mathbf{X}_t^{\tau_{k+1}}
    =
    \mathbf{X}_t^{\tau_k}
    + \Delta\tau_k\,
    v_\theta(\mathbf{X}_t^{\tau_k},\tau_k;\mathcal{H}_t).
    \label{eq:flow-matching-euler}
\end{equation}

Some implementations parameterize the same path by the remaining noise level
$\sigma=1-\tau$, which is integrated from $1$ to $0$. Defining the
corresponding field as
$\tilde v_\theta=-v_\theta$, the clean endpoint estimate under the linear path is
\begin{equation}
    \hat{\mathbf{X}}_t^1
    =
    \mathbf{X}_t^\sigma
    -\sigma\,
    \tilde v_\theta(\mathbf{X}_t^\sigma,\sigma;\mathcal{H}_t).
    \label{eq:clean-endpoint-estimate}
\end{equation}
We use superscripts exclusively for flow time or noise level and subscripts for
environment time. This separation is important for FBFM, which modifies the
inference-time vector field while preserving the pretrained Flow-Matching model.

For quick reference, Appendix~\ref{app:notation-index} provides a consolidated
index of the notation used throughout the paper.


\section{Method}
\label{sec:method}


\subsection{Feedback Flow Matching with Overlapping Chunks}
\label{sec:method:overlap}

FBFM turns chunked WAM inference into a feedback process without modifying or
retraining the pretrained model. Suppose that a new chunk is generated at
environment time $t$ while its preceding action chunk is still being executed.
After both chunks are aligned to global environment time, let
\begin{equation}
    \mathcal I_t^A
    =
    \{\,i:a_{t+i}\in\mathrm{overlap}\,\}.
    \label{eq:action-overlap-set}
\end{equation}
denote their overlap. For every $i\in\mathcal I_t^A$, the action inherited from
the preceding chunk, $a_{t+i}^{\mathrm{prev}}$, is a committed target for the
corresponding slot of the new chunk. The target covers the full aligned overlap,
independently of the current execution pointer.

Execution and generation proceed concurrently. Executing
$a_{t+i-1}^{\mathrm{prev}}$ produces a real transition whose observation is encoded
as $z_{t+i}$. At solver evaluation $k$, the available feedback is
\begin{equation}
    \mathcal F_{t,k}
    =
    \left\{
    (i,z_{t+i}): z_{t+i}\text{ arrives before }k
    \right\}.
    \label{eq:dynamic-feedback-set}
\end{equation}

The solver-start history $\mathcal H_t$ remains fixed; newly arriving states are
represented explicitly by $\mathcal F_{t,k}$ and its dynamic mask. Thus real
transitions progressively constrain the active chunk without redefining the
pretrained model's native condition.

\begin{figure*}[t]
    \centering
    \includegraphics[width=0.82\textwidth]{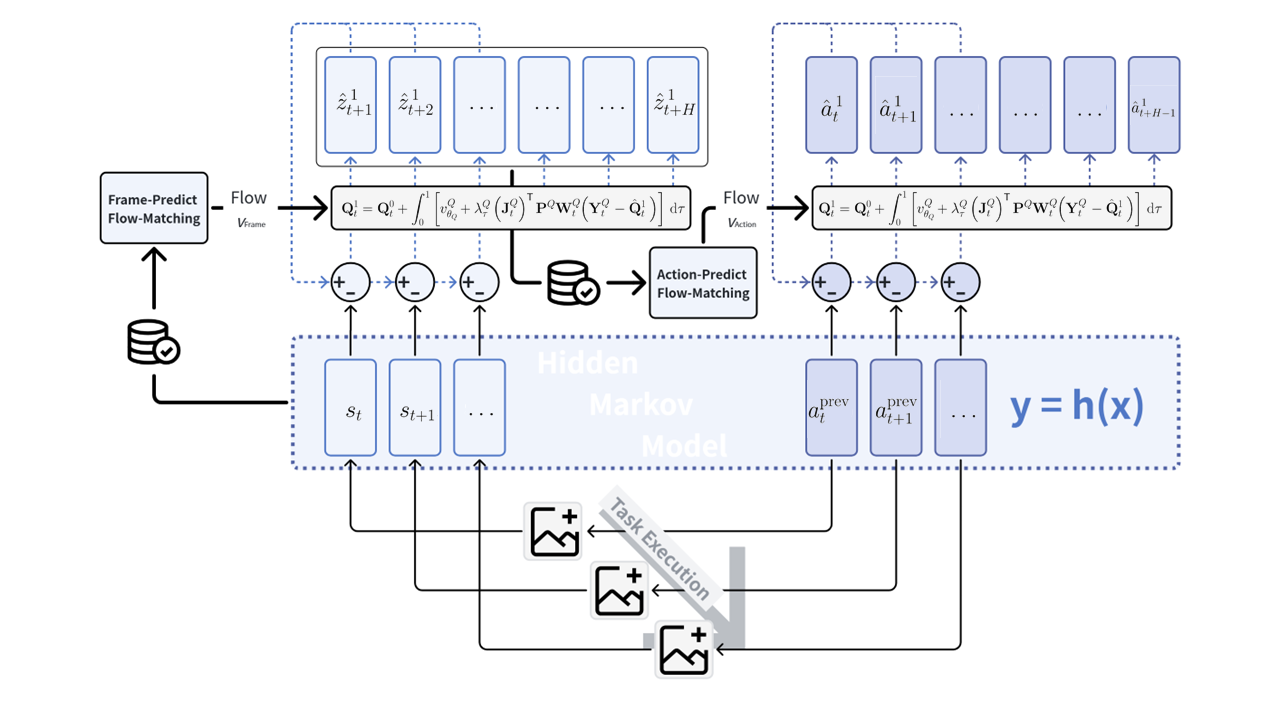}
    \caption{FBFM for a stage-wise WAM. While the preceding chunk is executed,
    encoded real observations progressively constrain the state flow. The latest
    corrected state context conditions the action flow, whose overlap is constrained
    by the committed actions inherited from the preceding chunk.}
    \label{fig:fbfm-stage-wise}
\end{figure*}

\subsection{Unified Pseudoinverse-Guided Flow Correction}
\label{sec:method:unified-guidance}

FBFM adapts pseudoinverse guidance~\cite{song2023pseudoinverse} to the active
flow of a WAM. Let $Q\in\{Z,A,X\}$ indicate whether the generated variable
$\mathbf Q_t\in\mathbb R^{D_Q}$ is a latent-state chunk, an action chunk, or
their joint representation. The corresponding frozen vector field, parameters,
flow time, and native conditioning context are denoted by
$v_{\theta_Q}^Q$, $\theta_Q$, $\tau_k^Q$, and
$\mathcal K_{t,k}^Q$, respectively.

Following the inverse-problem view, a clean variable produces a feedback
measurement
\begin{equation}
    \mathbf m_{t,k}^Q
    =h_Q(\mathbf Q_t^1)+\boldsymbol\eta,
    \qquad
    \boldsymbol\eta\sim\mathcal N(\mathbf 0,\sigma_y^2\mathbf I),
    \label{eq:fbfm-measurement}
\end{equation}
where $h_Q$ maps the generation space to the feedback space and
$h_Q^\dagger$ is its generalized inverse. We interpret them as a
Moore--Penrose-compatible encoder--decoder pair and store the lifted target as
$\mathbf Y_{t,k}^Q:=h_Q^\dagger(\mathbf m_{t,k}^Q)$. Pseudoinverse guidance
compares this target with
$h_Q^\dagger(h_Q(\hat{\mathbf Q}_{t,k}^1))$. Because FBFM feedback and model
predictions are already represented in aligned latent or action coordinates, we
use
$h_Q^\dagger(h_Q(\hat{\mathbf Q}_{t,k}^1))\approx
\hat{\mathbf Q}_{t,k}^1$ on the selected subspace. Appendix~\ref{app:pseudoinverse}
states the corresponding consistency assumptions.

Let $\mathbf W_{t,k}^Q$ select or confidence-weight the constrained
coordinates, and let the separate operator $\mathbf P^Q$ balance their scale.
For compactness, write
$\bar v_{t,k}^Q
=v_{\theta_Q}^Q(\mathbf Q_t^{\tau_k^Q},\tau_k^Q;\mathcal K_{t,k}^Q)$.
At every solver evaluation, all FBFM instantiations use the same guided field:
\begingroup
\small
\begin{equation}
\begin{aligned}
v_{\mathrm{FBFM}}^Q
&=\bar v_{t,k}^Q+\lambda_{\tau_k^Q}^Q
(\mathbf J_{t,k}^Q)^{\mathsf T}\mathbf P^Q\mathbf W_{t,k}^Q
\!\left(\mathbf Y_{t,k}^Q-\hat{\mathbf Q}_{t,k}^1\right),\\
\hat{\mathbf Q}_{t,k}^1
&=\mathbf Q_t^{\tau_k^Q}+(1-\tau_k^Q)\bar v_{t,k}^Q,\quad
\mathbf J_{t,k}^Q=\partial\hat{\mathbf Q}_{t,k}^1/
\partial\mathbf Q_t^{\tau_k^Q}.
\end{aligned}
\label{eq:fbfm-guided-field}
\end{equation}
\endgroup
Here, $\lambda_{\tau_k^Q}^Q$ is the flow-time-dependent guidance strength.
This formulation follows the Flow-Matching inpainting construction of
Black et al.~\cite{black2025realtime}, while making the support mask and
modality preconditioner explicit. The remainder of this section only specifies
the assignments of
$(\mathbf Q,\theta_Q,\mathcal K,\mathbf Y,\mathbf W,\mathbf P)$ for stage-wise
and joint-generation WAMs.

\subsection{FBFM for Stage-Wise World--Action Models}
\label{sec:method:stage-wise}

A stage-wise WAM factorizes latent-state and action generation as
\begin{equation}
    p_\theta(\mathbf Z_t,\mathbf A_t\mid\mathcal H_t)
    =
    p_{\theta_Z}(\mathbf Z_t\mid\mathcal H_t)
    p_{\theta_A}(\mathbf A_t\mid\mathbf Z_t,\mathcal H_t),
    \label{eq:stage-wise-factorization}
\end{equation}
and realizes the two factors with separate vector fields $v_{\theta_Z}^Z$ and
$v_{\theta_A}^A$. FBFM therefore applies separate state and action corrections,
while passing the corrected state representation to the action stage.

The stage-wise procedure preserves the WAM's state-first, action-second order.
While the preceding chunk is executed, each overlap action
$a_{t+i-1}^{\mathrm{prev}}$ induces a transition in the physical or simulated
environment according to $P(\cdot\mid s_{t+i-1},a_{t+i-1}^{\mathrm{prev}})$. The
resulting state is sensed and encoded as $z_{t+i}$, which activates the
corresponding mask and constrains the ongoing state flow. The subsequent action
flow is conditioned on the latest corrected state context and is directly
constrained by the committed actions in the cross-chunk overlap. State feedback
that arrives after the state flow has terminated refreshes the context used by the
ongoing action flow rather than restarting state generation.

The left side of Figure~\ref{fig:masked-guidance-comparison} makes the resulting
separation explicit: the two stages use distinct endpoint Jacobians and distinct
masks, even though feedback collected during execution connects them temporally.

\subsubsection{State Flow: Dynamic State-Feedback Guidance}

For the state stage, the unified correction uses
$Q=Z$, $\theta_Q=\theta_Z$,
$\mathcal K_{t,k}^Z=\mathcal H_t$, and
$\mathbf P^Z=P_Z\mathbf I_{D_Z}$. At state-solver flow time
$\tau_k^Z$, we encode $\mathcal F_{t,k}$ as an aligned state-feedback target
$\mathbf Y_{t,k}^Z\in\mathbb R^{D_Z}$ and a block-diagonal mask
\begin{equation}
    \begin{aligned}
    \mathbf W_{t,k}^Z
    &=
    \operatorname{Diag}
    \!\left(w_{t,1}^{Z,k},\ldots,w_{t,H}^{Z,k}\right)
    \otimes\mathbf I_{d_z},\\
    w_{t,i}^{Z,k}>0
    &\Longleftrightarrow
    (i,z_{t+i})\in\mathcal F_{t,k}.
    \end{aligned}
    \label{eq:dynamic-state-mask}
\end{equation}
For an observed slot, the corresponding block of $\mathbf Y_{t,k}^Z$ is
$z_{t+i}$; unobserved blocks may be filled arbitrarily because their weights are
zero. Binary weights impose hard state inpainting, while weights in $[0,1]$
permit confidence weighting. Refreshing $(\mathcal F_{t,k},\mathbf Y_{t,k}^Z,
\mathbf W_{t,k}^Z)$ before each solver evaluation makes newly available states
affect the remaining steps of the same active chunk through
Eq.~\eqref{eq:fbfm-guided-field}.

\subsubsection{Action Flow: State-Context Refresh and Previous-Action Consistency}

\paragraph{State-context refresh.}

After the state flow terminates, let $\hat{\mathbf Z}_t$ be its generated endpoint.
Additional state feedback may arrive while the action flow is still running. At an
action-solver evaluation $k$, we reuse $\mathbf Y_{t,k}^Z$ and
$\mathbf W_{t,k}^Z$ for the latest feedback snapshot available at that evaluation
and form the corrected state representation by
\begin{equation}
    \begin{aligned}
    \check{\mathbf Z}_{t,k}
    &=\left(\mathbf I_{D_Z}-\mathbf W_{t,k}^Z\right)\hat{\mathbf Z}_t
      +\mathbf W_{t,k}^Z\mathbf Y_{t,k}^Z,\\
    \mathcal C_{t,k}^Z
    &=\Phi_Z(\check{\mathbf Z}_{t,k},\mathcal H_t).
    \end{aligned}
    \label{eq:state-context-refresh}
\end{equation}
Here, $\Phi_Z$ denotes the native context-construction interface of the stage-wise
WAM. It may be realized by latent tokens, intermediate features, or an updated
attention memory; FBFM does not prescribe its storage mechanism.

\paragraph{Previous-action consistency.}

Let $\mathbf Y_t^A\in\mathbb R^{D_A}$ contain the actions
$a_{t+i}^{\mathrm{prev}}$ aligned to the new action horizon, with arbitrary values
outside $\mathcal I_t^A$. We represent the general previous-action constraint by a
nonnegative weighting operator
\begin{equation}
    \mathbf W_t^A\in[0,1]^{D_A\times D_A},
    \qquad D_A=H d_a,
    \label{eq:general-action-weight}
\end{equation}
whose support lies on the aligned overlap. A common hard-overlap specialization is
\begin{equation}
    \mathbf W_t^A
    =
    \operatorname{Diag}
    \!\left(
        \mathbb 1[0\in\mathcal I_t^A],\ldots,
        \mathbb 1[H-1\in\mathcal I_t^A]
    \right)
    \otimes\mathbf I_{d_a}.
    \label{eq:hard-overlap-action-mask}
\end{equation}

Crucially, $\mathbf Y_t^A$ and $\mathbf W_t^A$ describe cross-chunk consistency,
not the current execution pointer. They therefore remain fixed throughout generation
of the new chunk, regardless of which committed overlap actions have already been
executed.

For the action stage, the unified correction uses
$Q=A$, $\theta_Q=\theta_A$,
$\mathcal K_{t,k}^A=(\mathcal H_t,\mathcal C_{t,k}^Z)$, and
$\mathbf P^A=P_A\mathbf I_{D_A}$, with the fixed assignments
$\mathbf Y_{t,k}^A=\mathbf Y_t^A$ and
$\mathbf W_{t,k}^A=\mathbf W_t^A$. Substituting them into
Eq.~\eqref{eq:fbfm-guided-field} makes the
latest refreshed state context condition the endpoint predictor while the overlap
target directly constrains the action flow.

Thus, asynchronous state feedback updates the state flow and action context,
whereas the overlap target remains fixed and directly constrains the action flow.

\subsection{FBFM for Joint-Generation World--Action Models}
\label{sec:method:joint}

A joint-generation WAM models the latent state and action chunks with a single
conditional distribution,
\begin{equation}
    p_\theta(\mathbf X_t\mid\mathcal H_t),
    \qquad
    \mathbf X_t
    =
    \begin{bmatrix}
        \mathbf Z_t\\
        \mathbf A_t
    \end{bmatrix}.
    \label{eq:joint-generation-model}
\end{equation}
Unlike a stage-wise WAM, it transports both modalities with one vector field
$v_\theta^X$ and does not require a separate state-to-action context handoff. The
dynamic state feedback and the fixed previous-action target are instead assembled
into one joint constraint and applied at every evaluation of the same Flow-Matching
solver.

\begin{figure*}[t]
    \centering
    \includegraphics[width=0.82\textwidth]{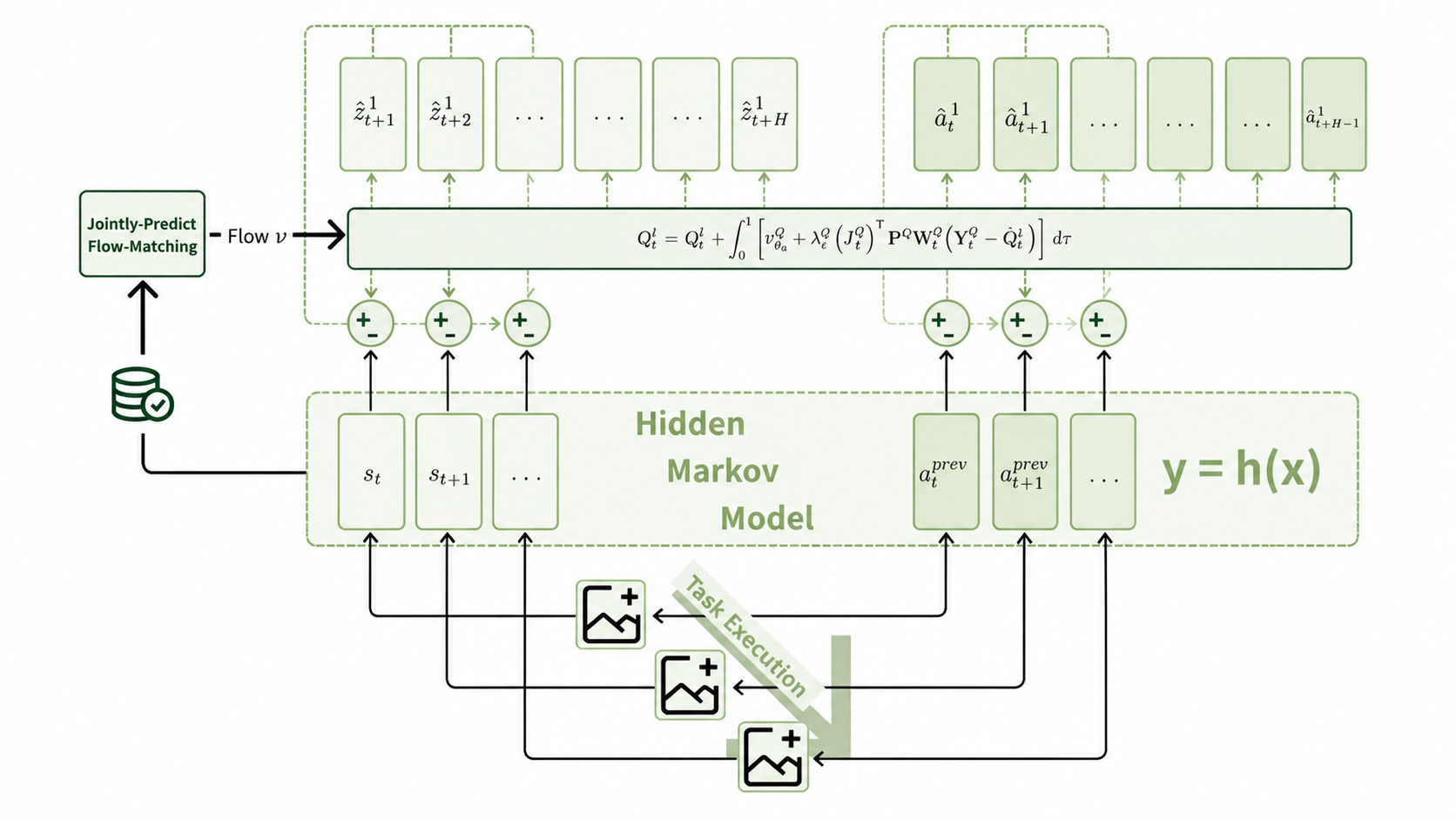}
    \caption{FBFM for a joint-generation WAM. Encoded transitions observed while
    the preceding chunk is executed and committed actions from the cross-chunk
    overlap jointly constrain a single state-action flow. Through the cross-modal
    blocks of the endpoint Jacobian, state feedback can directly correct the action
    coordinates of this flow.}
    \label{fig:fbfm-joint}
\end{figure*}

\subsubsection{Joint State--Action Guidance}

Using the state-feedback quantities from Section~\ref{sec:method:stage-wise} and
the same aligned previous-action target, we define
\begin{equation}
    \mathbf Y_{t,k}^X
    =
    \begin{bmatrix}
        \mathbf Y_{t,k}^Z\\
        \mathbf Y_t^A
    \end{bmatrix},
    \label{eq:joint-feedback-target}
\end{equation}
and
\begin{equation}
    \begin{aligned}
    \mathbf W_{t,k}^X
    &=
    \begin{bmatrix}
        \mathbf W_{t,k}^Z & \mathbf 0\\
        \mathbf 0 & \mathbf W_t^A
    \end{bmatrix},\\
    D_X&=D_Z+D_A,\qquad D_Z=H d_z.
    \end{aligned}
    \label{eq:joint-feedback-mask}
\end{equation}

The corresponding modality preconditioner is
\begin{equation}
    \mathbf P^X
    =
    \operatorname{blkdiag}
    \!\left(P_Z\mathbf I_{D_Z},P_A\mathbf I_{D_A}\right).
    \label{eq:joint-preconditioner}
\end{equation}

The state block of $\mathbf W_{t,k}^X$ is refreshed whenever a new
$z_{t+i}$ becomes available, whereas its action block remains fixed by the aligned
cross-chunk overlap. For joint generation, the remaining unified assignments are
$Q=X$, $\theta_Q=\theta$, and $\mathcal K_{t,k}^X=\mathcal H_t$.
Substituting these quantities into
Eq.~\eqref{eq:fbfm-guided-field} gives the
complete joint update. One guidance evaluation simultaneously enforces observed state slots and
the committed action overlap while leaving all unobserved and unconstrained
coordinates to the pretrained joint model.

\begin{figure*}[t]
    \centering
    \begin{minipage}[t]{0.49\textwidth}
        \centering
        \includegraphics[width=\linewidth]{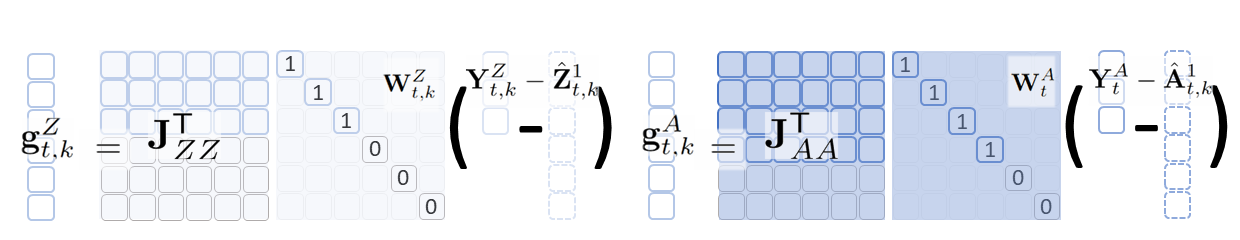}
        \vspace{0.2em}

        \includegraphics[width=\linewidth]{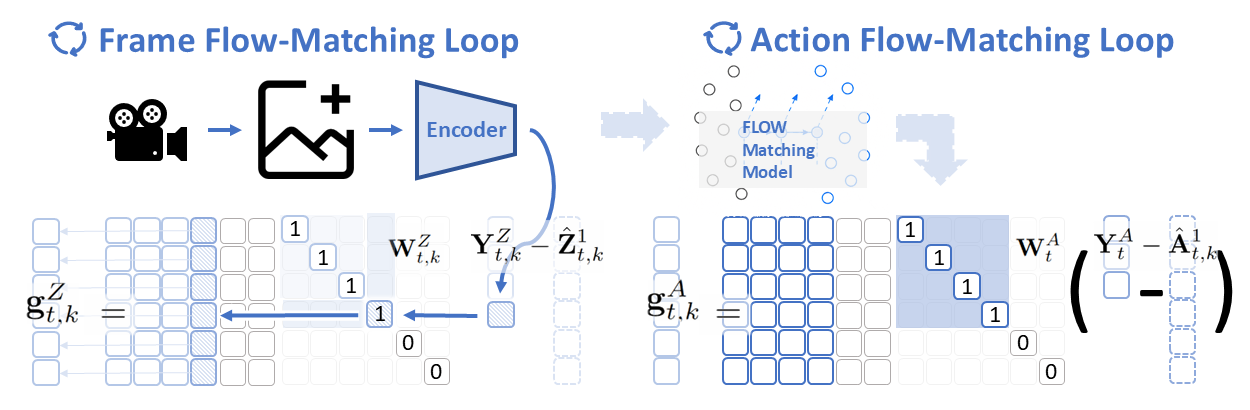}
    \end{minipage}
    \hfill
    \begin{minipage}[t]{0.49\textwidth}
        \centering
        \includegraphics[width=\linewidth]{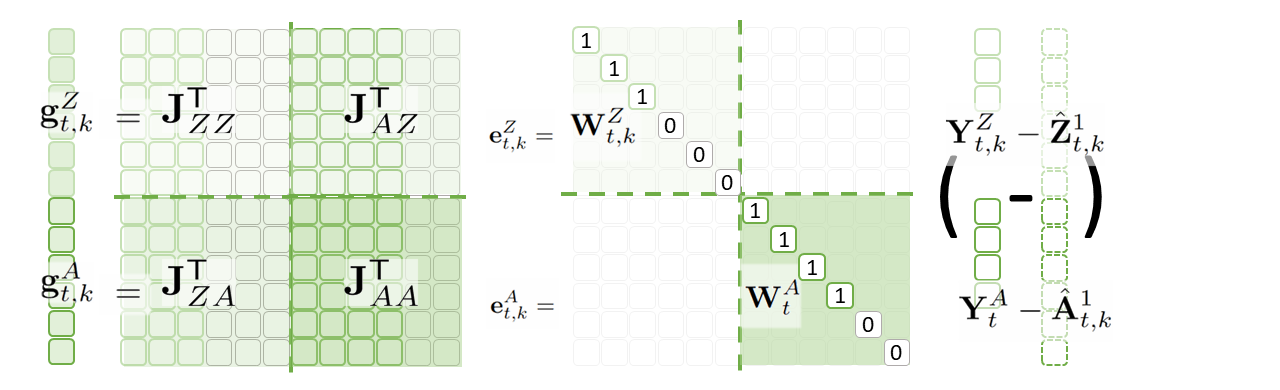}
        \vspace{0.2em}

        \includegraphics[width=\linewidth]{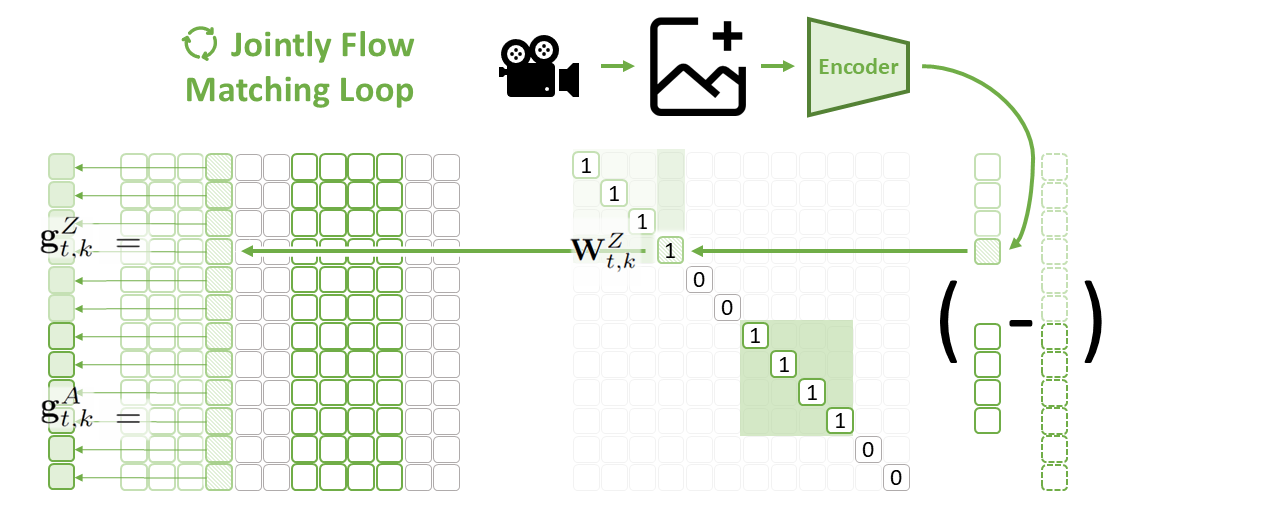}
    \end{minipage}
    \caption{Masked FBFM guidance in stage-wise and joint-generation WAMs.
    \emph{Left:} stage-wise WAMs use distinct endpoint Jacobians and masks for
    the state and action flows; real state feedback activates dynamic state-mask
    entries, while the action-overlap mask remains fixed.
    \emph{Right:} joint-generation WAMs form a block-diagonal state-action mask,
    and the full endpoint-Jacobian transpose propagates corrections across state
    and action coordinates.}
    \label{fig:masked-guidance-comparison}
\end{figure*}

Figure~\ref{fig:masked-guidance-comparison} illustrates both parts of this joint
update: the mask forms modality-specific discrepancies, while the transpose of the
full endpoint Jacobian propagates them across state and action coordinates.

\subsubsection{Direct State-to-Action Correction}

For $Q,R\in\{Z,A\}$, define the endpoint-Jacobian block
$\mathbf J_{QR}:=\partial\hat{\mathbf Q}_{t,k}^1/
\partial\mathbf R_t^{\tau_k^X}$. Partitioning
$\mathbf e_{t,k}^X=[\mathbf e_{t,k}^Z;\mathbf e_{t,k}^A]$ and
$\mathbf g_{t,k}^X=[\mathbf g_{t,k}^Z;\mathbf g_{t,k}^A]$ gives
\begin{equation}
    \begin{aligned}
    \mathbf g_{t,k}^Z
    &=\mathbf J_{ZZ}^{\mathsf T}\mathbf e_{t,k}^Z
      +\mathbf J_{AZ}^{\mathsf T}\mathbf e_{t,k}^A,\\
    \mathbf g_{t,k}^A
    &=\mathbf J_{ZA}^{\mathsf T}\mathbf e_{t,k}^Z
      +\mathbf J_{AA}^{\mathsf T}\mathbf e_{t,k}^A,\\
    \mathbf e_{t,k}^A=\mathbf 0
    \quad\Longrightarrow\quad
    \mathbf g_{t,k}^A
    &=\mathbf J_{ZA}^{\mathsf T}\mathbf e_{t,k}^Z
      =\left(\frac{\partial\hat{\mathbf Z}_{t,k}^1}
      {\partial\mathbf A_t^{\tau_k^X}}\right)^{\mathsf T}
      \mathbf e_{t,k}^Z.
    \end{aligned}
    \label{eq:direct-state-to-action-correction}
\end{equation}

Whenever the learned joint endpoint predictor couples states and actions,
$\mathbf J_{ZA}\neq\mathbf 0$, and a discrepancy in an observed state slot
therefore produces a nonzero correction on the action coordinates in the same
solver step. This is the principal distinction from the stage-wise formulation:
state feedback does not influence action generation only through a subsequently
refreshed context; it can directly modify the action flow through the cross-modal
Jacobian of the joint predictor. The reciprocal cross block also allows an action
residual to affect state coordinates, although FBFM primarily uses the former path
to inject real state transitions into action generation.

\subsection{Training-Free Inference}
\label{sec:method:training-free}

FBFM changes only the inference-time vector fields and context supplied to the
pretrained WAM. The joint parameter set $\theta$, or the stage-wise parameter sets
$\theta_Z$ and $\theta_A$, remain frozen, and the VJPs are obtained by automatic
differentiation through the clean-endpoint predictors.
The method does not require a fixed ratio between environment steps and solver
steps. It only requires feedback available before a solver evaluation to be visible
to that evaluation. The concrete scheduling mechanism is therefore an experimental
property rather than part of the method definition.


\section{Experiments}
\label{sec:experiments}


We evaluate FBFM in two complementary tracks that cover the two WAM generation
factorizations considered in Section~\ref{sec:method}. Comparisons are made
within each track using the same frozen base model; absolute scores are not
compared across architectures or benchmarks.

\subsection{Models, Benchmarks, and Tasks}
\label{subsec:models-benchmarks-tasks}

\paragraph{Models.}
We instantiate FBFM on LingBot-VA~\cite{li2026lingbot} and
DreamZero~\cite{ye2026dreamzero}. The LingBot-VA track uses the official
checkpoint post-trained on RoboTwin2.0. Its released inference implementation
provides the stage-wise WAM instance, in which the state trajectory is
generated before the action trajectory. The DreamZero track uses a
LIBERO-post-trained checkpoint obtained at SFT step 26,000 with the RLinf
DreamZero SFT recipe~\cite{yu2025rlinf}.
DreamZero jointly generates future states and actions and therefore provides
the joint-generation WAM instance.

\paragraph{Benchmarks and tasks.}
We evaluate the LingBot-VA track on RoboTwin2.0~\cite{chen2025robotwin2} and the
DreamZero track on LIBERO~\cite{liu2023libero}. The DreamZero comparison covers
all four standard suites---LIBERO-Spatial, LIBERO-Object,
LIBERO-Goal, and LIBERO-10---with 10 tasks per suite. Each task is evaluated
once at each reset ID from 0 to 19, giving 20 episodes per task and 800 episodes
per method.
Both benchmarks provide environment-side, task-specific completion predicates.
We use these predicates without manual relabeling and report the fraction of
successful episodes as the task success rate.

\subsection{FBFM Instantiations}
\label{subsec:fbfm-instantiations}

For both tracks, FBFM is inserted only at inference time. We retain the
checkpoint, native solver and cache schedule, classifier-free guidance, and
chunk dimensions of each WAM, with all pretrained parameters frozen. A
deterministic pseudo-asynchronous clock couples environment transitions to
solver evaluations, controlling feedback timing independently of wall-clock
latency. Exact tensor layouts, solver schedules, feedback-release rules, and
guidance parameters are reported in Appendix~\ref{app:implementation}.

\paragraph{Stage-wise generation: LingBot-VA.}
We preserve the released video-first, action-second inference order. While the
preceding action suffix is executed, frozen-VAE observations update
time-aligned constraints in the active video flow. The corrected video context
then conditions action generation, while the same preceding suffix provides a
fixed action-prefix constraint.

\paragraph{LingBot-VA mechanism diagnostic.}
We isolate the two computation-level links induced by state-context refresh
using four paired task--trial units from
\texttt{adjust\_bottle} and \texttt{pick\_diverse\_bottles} under the
RoboTwin2.0 randomized setting. First, we compare the wave-0 predicted next-state latent
with the next realized latent under matched RTC~\cite{black2025realtime} and
FBFM initial conditions. Second, a CacheCut intervention switches only the
installed RTC versus FBFM cache while holding history, noise, constraints, and
the solver schedule fixed; repeating the same RTC cache estimates the numerical
floor. This diagnostic tests state-feedback and cache-to-action influence, not
task-success differences.

\paragraph{Joint generation: DreamZero.}
We retain DreamZero's joint state--action solver. Guidance is recomputed at all
solver updates while respecting the model's native DiT cache schedule. At
cached updates, the latest native velocity and endpoint Jacobian are reused,
but the residual and guided field are formed from the current solver sample.
Joint differentiation therefore preserves the cross-modal Jacobian path from
state feedback to action coordinates. Observations are retained causally and
installed only when a complete latent target is available at the checkpoint's
video stride; the preceding actions remain a fixed prefix target.

The implementation exposes a common mask-controlled path for later controlled
comparisons. Within each architecture, the unguided, action-only, and full FBFM
modes share the checkpoint, noise, solver budget, and pseudo-clock, differing
only in their active masks. DreamZero's native synchronous rollout is recorded
separately.

\subsection{Real-World Observation Prediction}
\label{subsec:real-world-observation-prediction}

Finally, we evaluate state feedback on an RGB sequence captured during a
physical robot-arm ball-stopping trial. A frozen Wan2.2-TI2V-5B
model~\cite{wan2025} receives the same image anchor, prompt, seed, and 50-step
solver in both conditions: the Base performs native video prediction, whereas
FBFM causally encodes the next 120 RealSense D435i frames into 30 latent
measurements released across the active solve. This recorded-execution
diagnostic evaluates real-world visual-state prediction rather than task
success; full-resolution videos and preprocessing metadata are included in the
supplementary material.

\begin{figure*}[!t]
    \centering
    \includegraphics[width=0.82\textwidth]{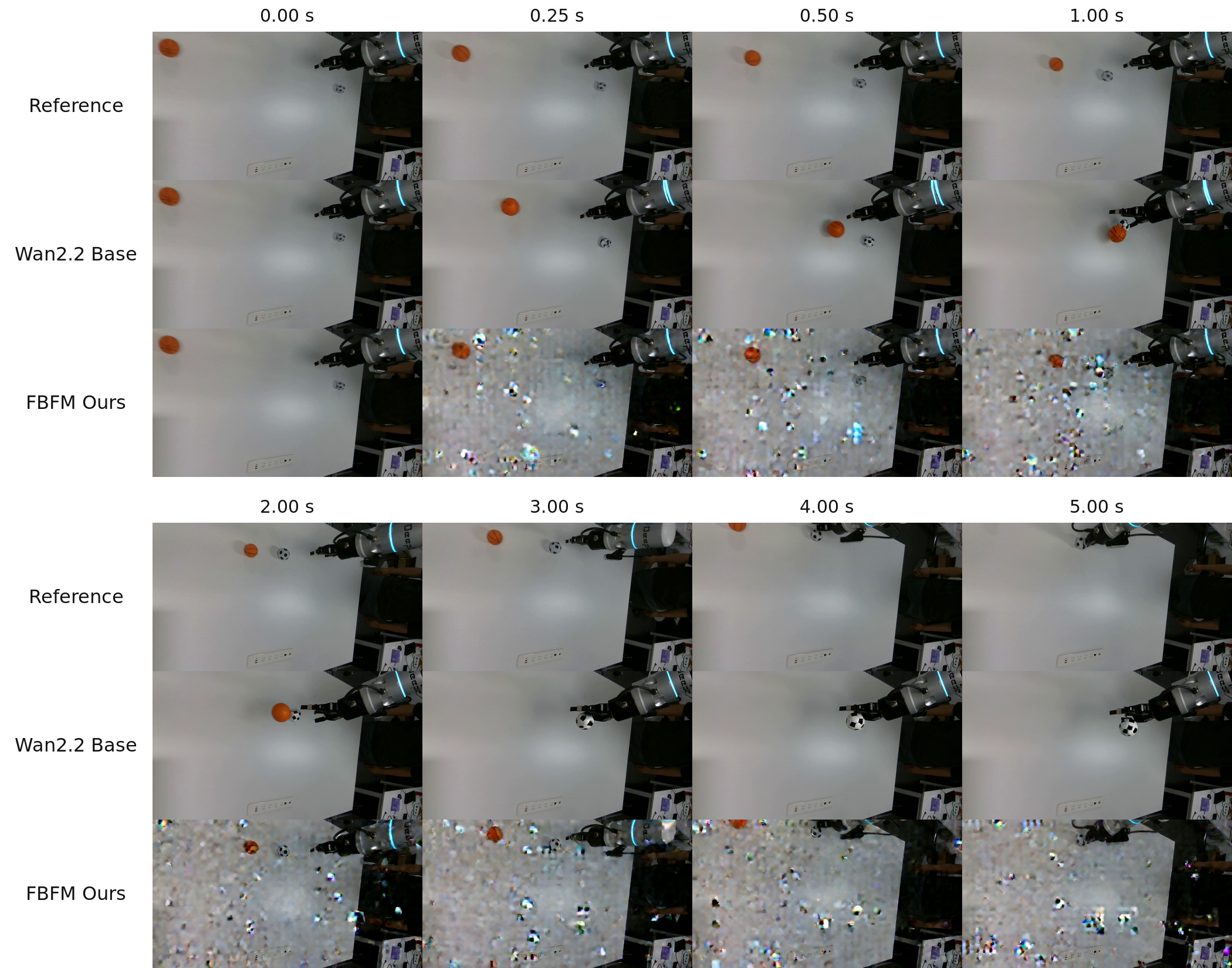}
    \caption{Real-world robot-arm ball-stopping observation prediction. The RGB
    sequence was recorded from a physical robot task with a RealSense D435i.
    The upper block shows 0, 0.25, 0.5, and 1~s, and the lower block shows 2,
    3, 4, and 5~s. Within each block, rows show the recorded reference, Wan2.2
    Base without feedback, and FBFM using all 30 measured latent slots.}
    \label{fig:wan-robot-arm-ball-stop}
\end{figure*}


\section{Results and Discussion}
\label{sec:results-discussion}


\subsection{LingBot-VA on RoboTwin2.0}

We compare the frozen LingBot-VA checkpoint with no inference-time feedback
(Base) against the same checkpoint equipped with FBFM (Ours) on all 42 selected
tasks under both Clean and Randomized configurations. Because the available
task cells contain either 10 or 20 episodes, we first compute each task-cell
success rate and then macro-average them, giving every task equal weight. CPU
and GPU rendering are treated as equivalent benchmark backends; their source
provenance and the original integer counts are retained in
Appendix~\ref{app:detailed-results}.

\begin{table}[t]
    \centering
    \small
    \setlength{\tabcolsep}{4pt}
    \begin{tabular}{lccc}
        \toprule
        Method & Clean & Rand. & Overall \\
        \midrule
        Base & 80.5 & 79.8 & 80.1 \\
        FBFM & \textbf{83.3} $\uparrow$ & \textbf{82.9} $\uparrow$ &
        \textbf{83.1} $\uparrow$ \\
        \bottomrule
    \end{tabular}
    \caption{LingBot-VA success rates (\%). Each entry is an equal-weight
    macro average over task-configuration cells.}
    \label{tab:lingbot-robotwin-summary}
\end{table}

FBFM improves Clean and Randomized success rates by 2.86 and 3.10 percentage
points, respectively, for an overall gain of 2.98 points. It improves 19 of 42
Clean task cells and 15 of 42 Randomized cells, while matching Base on 14 and 20
cells, respectively. With the checkpoint, task predicates, solver budget, and
inference configuration otherwise fixed, these broad task-level gains support
the effectiveness of inference-time feedback in stage-wise WAM execution.

\subsection{LingBot-VA Mechanism Analysis}

\begin{figure}[H]
    \centering
    \includegraphics[width=\columnwidth]{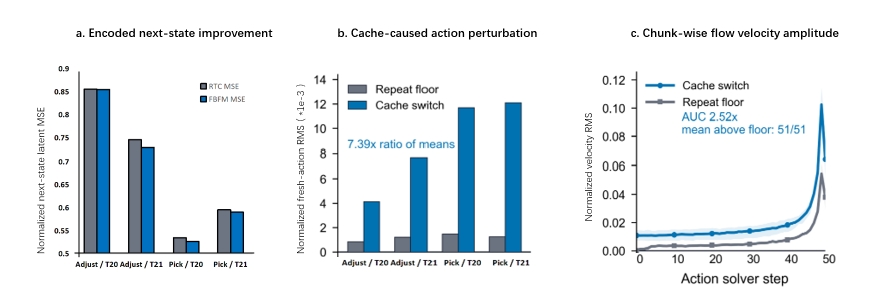}
    \caption{LingBot-VA mechanism diagnostic. (a) Paired wave-0 next-state
    latent MSE. (b) Fresh-action RMS under a same-cache repeat and an RTC--FBFM
    cache switch. (c) Cache-switch and repeat-floor velocity RMS across the 51
    action-solver steps.}
    \label{fig:lbva-aux-mechanism}
\end{figure}

Across four paired task--trial units, FBFM lowers wave-0 next-state latent MSE
from 0.6828 to 0.6751 (1.13\%). Switching only the installed RTC/FBFM cache
changes normalized fresh-action RMS by 0.00890, 7.39$\times$ the repeat floor;
velocity RMS stays above that floor for all 51 steps, with a 2.52$\times$ AUC
ratio. This traces the intended state-feedback-to-action path.

\subsection{DreamZero on LIBERO}

We compare the frozen DreamZero checkpoint with and without FBFM on
LIBERO-Spatial, LIBERO-Object, LIBERO-Goal, and LIBERO-10; LIBERO-90 is
excluded. Both evaluations use the same tasks, 20 official initial states per
task, 480-step horizon, and delayed pseudo-asynchronous overlap protocol.
DreamZero's native synchronous rollout is not used as the Base. For concision,
the main table reports all four evaluated suites and their pooled total.
Appendix~\ref{app:detailed-results} reports every task.

\begin{table}[t]
    \centering
    \small
    \setlength{\tabcolsep}{4pt}
    \begin{tabular}{lccccc}
        \toprule
        Method & Spatial & Object & Goal & L-10 & Total \\
        \midrule
        Base & 78.5 & 73.0 & 68.5 & 60.5 & 70.1 \\
        FBFM & 77.0 & 72.0 &
        \textbf{71.0} $\uparrow$ & \textbf{63.0} $\uparrow$ &
        \textbf{70.8} $\uparrow$ \\
        \bottomrule
    \end{tabular}
    \caption{DreamZero success rates (\%). Total pools all four evaluated
    suites; up-arrows mark improvements over the corresponding Base result.}
    \label{tab:dreamzero-libero-summary}
\end{table}

FBFM improves both LIBERO-Goal and LIBERO-10 by 2.5 percentage points, while
LIBERO-Spatial and LIBERO-Object decrease by 1.5 and 1.0 points, respectively.
Across all 800 episodes, the pooled success rate increases by 0.625 points. We
therefore interpret the current result as a modest aggregate gain with
heterogeneous suite-level effects rather than a uniform improvement. A likely
reason is DreamZero's accelerated inference design: one native model evaluation
is reused across multiple solver updates. FBFM recomputes its residual from the
current $\mathbf Z$ and $\mathbf A$, but reuses the latest native velocity and
endpoint Jacobian between DiT evaluations; this amortization can damp or delay
new corrections. Joint feedback is also sensitive to the relative scale of its
state and action residuals. We therefore use modality preconditioning and a
separately calibrated proportional gain. Appendix~\ref{app:dreamzero-gain-search}
contains the closed-loop motivation, configuration search, and task-level
sensitivity analysis; the selected value is an operational setting rather than
a universal optimum.

\subsection{Real-World Robot Observation Prediction}
\label{sec:discussion:real-video}

Figure~\ref{fig:wan-robot-arm-ball-stop} reports the recorded physical
robot-arm ball-stopping experiment described above. Both methods start from the
same image at the end of a two-second context and predict the same five-second,
121-frame horizon. The Base receives no later image, whereas FBFM causally
incorporates all 120 measured future RealSense frames as 30 latent feedback
slots. Base preserves a coherent robot and tabletop but departs from the
recorded ball evolution. FBFM moves the full-frame prediction closer to the
reference (MAE 9.63 to 9.27 and PSNR 20.06 to 23.10~dB) and better preserves
task-relevant physical information such as the ball position over time, even
though visual artifacts remain. This shows that FBFM can use observations
recorded from a physical robot task to constrain the active frame chunk beyond
pure open-loop prediction. Appendix~\ref{app:wan-video-codec-diagnostic}
separately analyzes the behavior observed when state feedback covers only a
limited prefix of the generated video and treats the resulting large-area
artifacts as a Wan2.2-specific codec/backbone mismatch hypothesis.


\section{Related Works}
\label{sec:related-works}

Diffusion and Flow Matching established iterative transport as a general
interface for structured generation
\cite{sohl2015deep,ho2020denoising,song2021ddim,song2021score,rombach2022latent,
lipman2023flow,liu2023rectified}. Their intermediate sampling states also admit
inference-time constraints. RePaint conditions diffusion on known pixels, while
$\Pi$GDM and its continuous-flow extensions propagate measurement residuals
through a frozen generator by vector--Jacobian products
\cite{lugmayr2022repaint,song2023pseudoinverse,pokle2024training}. This
pseudoinverse-guided interface is the generative foundation of FBFM.

World models developed in parallel from learned visual rollouts and recurrent
latent dynamics toward task-oriented representations for imagined control
\cite{ha2018world,kaiser2020simple,hafner2019planet,hafner2020dreamer,
hafner2021dreamerv2,schrittwieser2020muzero,hansen2022tdmpc,hafner2025dreamerv3}.
Video diffusion, action-conditioned simulation, and interactive generation then
made explicit visual futures increasingly controllable
\cite{voleti2022mcvd,yang2024unisim,bruce2024genie,alonso2024diamond,zhu2025irasim}.
At their intersection, GR-1/GR-2 predict future images and actions, LingBot-VA
models video and action with observation refresh, and DreamZero jointly generates
both modalities \cite{wu2023gr1,cheang2024gr2,li2026lingbot,ye2026dreamzero}.
Fast-WAM instead removes explicit future imagination at inference
\cite{yuan2026fastwam}; FBFM addresses WAMs that retain this latent future and
must re-ground it during execution.

Generative decision models brought diffusion and inpainting from trajectory
planning to temporally correlated action chunks
\cite{janner2022diffuser,ajay2023decisiondiffuser,zhao2023act,
chi2023diffusionpolicy,ze2024dp3,liu2025rdt}. In parallel, RT-1/RT-2, Octo, and
OpenVLA scaled generalist robot policies through larger data and pretrained
vision--language representations
\cite{brohan2023rt1,brohan2023rt2,octo2024,kim2024openvla}. Flow-Matching
policies progressed from structured robot motion to generalist VLA control
\cite{braun2024rfmp,rouxel2024multisupport,zhang2025affordance}. The broad
empirical adoption of the Flow-Matching action expert in $\pi_0$ and
$\pi_{0.5}$ \cite{black2024pi0,pi2025pi05} is especially relevant: their
iterative velocity field enables RTC to impose a committed action prefix through
training-free inpainting or learned prefix conditioning
\cite{black2025realtime,black2025training}.

Asynchronous action-chunk execution must overlap inference with control while
maintaining cross-chunk continuity. Bidirectional Decoding selects coherent
candidate chunks; inference-time RTC directly constrains the active
Flow-Matching trajectory; its training-time counterpart, Legato, and PAINT
instead learn continuation or modify initialization
\cite{liu2025bid,black2025realtime,black2025training,liu2026legato,ho2026paint}.
These methods mainly constrain actions inherited from an earlier plan.

Other approaches react to information arriving during execution. RA-DP
interleaves denoising and action dequeueing, VLASH compensates inference delay
with predicted execution-time state, and A2C2/DCDP correct stale actions using
additional learned modules \cite{ye2025radp,tang2025vlash,sendai2025a2c2,
wu2026dcdp}. AsyncVLA and TIDAL likewise trade specialized training or a
lightweight controller for online action refinement
\cite{jiang2026asyncvla,sun2026tidal}. These methods complement FBFM but operate
primarily in action space.

Dynamics- and world-feedback methods provide the closest second line of work.
DynaGuide steers a diffusion policy through a separately trained latent dynamics
model \cite{du2025dynaguide}. Feedback World Model independently closes a
prediction--observation loop with a latent observer and action-aware guidance,
and provides an observer-error analysis \cite{an2026feedback}. AHA-WAM routes
current observations into reusable WAM context, while WA-LQR controls internal
WAM activations \cite{cai2026ahawam,hong2026walqr}. FBFM instead uses the frozen
WAM's own differentiable flow to impose time-aligned latent-state measurements
and committed actions within the actively generated multi-step chunk.


\section{Conclusion}
\label{sec:conclusion}

This paper studies a practical mismatch in world-action model execution:
long-horizon behavior needs continual grounding in real observations, while
existing chunk-wise refresh schemes only correct the model after a chunk has
already been generated. We proposed Feedback Flow Matching (FBFM), a
training-free inference mechanism that moves feedback into the active
flow-matching process. By expressing both committed actions and newly observed
latent states as masked pseudoinverse measurements, FBFM provides a common
correction interface for stage-wise and joint-generation WAMs.

Our experiments instantiate this idea on LingBot-VA on RoboTwin2.0 and
DreamZero on LIBERO, and further evaluate real-world observation prediction with
Wan2.2. The results show a 3.0-point task-configuration macro gain for the
stage-wise WAM, mechanism-level evidence that state feedback changes latent
prediction and propagates to action generation through the velocity field, and
modest but positive aggregate gains for the joint WAM after feedback-scale
tuning. The real-world video diagnostic
also indicates that FBFM can constrain physically meaningful visual evolution
beyond open-loop prediction. Together, these findings support FBFM as a simple
and extensible way to bridge open-loop generative execution with online
environment feedback.

Several limitations remain. The improvement on DreamZero is limited, likely
because its accelerated inference reuses model velocities and endpoint
Jacobians across multiple solver updates, which weakens the immediate effect of
new feedback. Our current implementation also uses the aligned-coordinate
approximation $h^\dagger(h(\hat{\mathbf X}))\approx\hat{\mathbf X}$ on the
mask-selected feedback subspace; this is practical for the latent and action
coordinates used here, but is not guaranteed for arbitrary nonlinear
encoder--decoder pairs. FBFM also adds Jacobian-vector-product and feedback-management
overhead at inference time; deploying it as a closed-loop robot policy requires
further engineering to recover real-time control frequency. Looking forward,
asynchronous feedback may make longer WAM chunks usable without abandoning
fine-grained correction. We also see two useful theoretical and algorithmic
directions: studying latent-space state transitions with nonlinear dynamics
tools, and replacing the current proportional gain with richer feedback
controllers, such as PID-style corrections, inside the flow-matching process.

\clearpage
\onecolumn
\raggedbottom
\appendix

\section{Notation Index}
\label{app:notation-index}

For quick reference, Table~\ref{tab:preliminaries-notation} indexes the core
notation introduced in Preliminaries, and Table~\ref{tab:method-notation} lists
the extensions used by the stage-wise and joint-generation FBFM formulations.

\begin{table}[H]
    \centering
    \small
    \renewcommand{\arraystretch}{1.12}
    \begin{tabular}{p{0.23\textwidth} p{0.71\textwidth}}
        \hline
        \textbf{Symbol} & \textbf{Definition}\\
        \hline
        $T,\mathcal{T},c_T,\rho_T$
        & A task, the task family, the condition associated with task $T$, and its initial-state distribution.\\
        $\mathcal{M},\mathcal{S},\mathcal{A},P$
        & The controlled Markov process, environment state space, action space, and transition kernel.\\
        $t,i,s_t$
        & Environment time, an offset within a chunk, and the physical environment state.\\
        $\mathcal{O},E,\mathcal{H}_t,\pi$
        & Sensor mapping, perceptual encoder, interaction history at time $t$, and the action-selection process.\\
        $H,d_a,d_z,D_A,D_Z,D_X,D$
        & Prediction horizon; dimensions of one action and one latent state; time-stacked action, state, and joint chunk dimensions, where $D_A=Hd_a$, $D_Z=Hd_z$, and $D_X=D_Z+D_A$; and a generic generation-space dimension.\\
        $a_t,\mathbf{A}_t$
        & A single-step action and its time-stacked action chunk vector.\\
        $z_t,\mathbf{Z}_t,\mathcal{Z}$
        & An encoded single-step latent state, its time-stacked future latent-state chunk vector, and the latent space.\\
        $x,\mathbf{X}_t$
        & A generic single-step variable and its chunk-level representation; instantiated as $a/\mathbf A$, $z/\mathbf Z$, or a joint state-action chunk.\\
        $\hat{\cdot},p_\theta,\theta$
        & A predicted or estimated quantity, the conditional WAM distribution, and its model parameters.\\
        $\tau,k,\tau_k,\Delta\tau_k,\sigma$
        & Continuous flow time, solver-step index, the corresponding flow-time point and integration step, and the remaining noise level $\sigma=1-\tau$.\\
        $\boldsymbol{\epsilon},\mathbf{0},\mathbf{I},\mathbf{X}_t^\tau$
        & Gaussian source noise, zero vector, identity matrix, and the intermediate chunk vector on the probability path.\\
        $\mathbf{u},v_\theta,\tilde v_\theta$
        & Conditional target velocity, learned Flow-Matching vector field, and its noise-level parameterization $\tilde v_\theta=-v_\theta$.\\
        $\mathcal{D},p(\tau),\mathcal{L}_{\mathrm{FM}}$
        & Training distribution, flow-time sampling distribution, and Flow-Matching objective.\\
        $\hat{\mathbf{X}}_t,\hat{\mathbf{X}}_t^1,f_\theta^\tau$
        & Generated chunk, predicted clean endpoint, and the endpoint predictor evaluated at flow time $\tau$.\\
        \hline
    \end{tabular}
    \caption{Core notation introduced in Preliminaries.}
    \label{tab:preliminaries-notation}
\end{table}

\begin{table}[H]
    \centering
    \small
    \renewcommand{\arraystretch}{1.12}
    \begin{tabular}{p{0.29\textwidth} p{0.65\textwidth}}
        \hline
        \textbf{Symbol} & \textbf{Definition}\\
        \hline
        $\mathcal I_t^A,a_{t+i}^{\mathrm{prev}}$
        & The action-slot overlap between the preceding and new chunks, and the preceding chunk's action aligned to an overlap slot.\\
        $\mathcal F_{t,k}$
        & Dynamic set of encoded real-state feedback available before solver evaluation $k$, kept separate from the solver-start history $\mathcal H_t$.\\
        $Q,D_Q,\mathbf Q_t$
        & Generic modality index $Q\in\{Z,A,X\}$, its generation-space dimension, and the corresponding latent-state, action, or joint chunk.\\
        $\theta_Q,v_{\theta_Q}^Q,\tau_k^Q,\mathcal K_{t,k}^Q$
        & Frozen parameters, vector field, flow time, and native conditioning context for modality $Q$.\\
        $\mathbf m_{t,k}^Q,h_Q,h_Q^\dagger,\boldsymbol\eta,\sigma_y$
        & Feedback-space measurement, feedback encoder and generalized lifting decoder, measurement noise, and its standard deviation.\\
        $\mathbf Y_{t,k}^Q,\mathbf W_{t,k}^Q$
        & Lifted feedback target and its support/confidence weighting operator.\\
        $\mathbf P^Q,P_Z,P_A$
        & Modality preconditioner and the state/action scale factors kept distinct from the support mask.\\
        $\hat{\mathbf Q}_{t,k}^1,f_{\theta_Q}^{Q,\tau_k^Q},\mathbf e_{t,k}^Q$
        & Predicted clean endpoint, its endpoint predictor, and the preconditioned masked discrepancy.\\
        $\mathbf J_{t,k}^Q,\mathbf g_{t,k}^Q,v_{\mathrm{FBFM}}^Q,\lambda_{\tau_k^Q}^Q$
        & Endpoint Jacobian, VJP correction, guided vector field, and flow-time-dependent guidance strength.\\
        $\theta_Z,\theta_A,v_{\theta_Z}^Z,v_{\theta_A}^A$
        & Frozen parameters and separate state/action vector fields of a stage-wise WAM.\\
        $\tau_k^Z,\tau_k^A,f_{\theta_Z}^{Z,\tau_k^Z},f_{\theta_A}^{A,\tau_k^A}$
        & State/action flow times and their clean-endpoint predictors at solver evaluation $k$.\\
        $\mathbf Y_{t,k}^Z,\mathbf W_{t,k}^Z,w_{t,i}^{Z,k}$
        & Aligned dynamic state-feedback target, its block mask/weighting operator, and the weight for state slot $i$.\\
        $\mathbf Y_t^A,\mathbf W_t^A$
        & Aligned committed-action target and the general previous-action weighting operator over the cross-chunk overlap.\\
        $\hat{\mathbf Z}_t,\check{\mathbf Z}_{t,k},\Phi_Z,\mathcal C_{t,k}^Z$
        & Generated state endpoint, its latest feedback-refreshed representation, the native state-context constructor, and the context supplied to the action flow.\\
        $\tau_k^X,v_\theta^X,f_\theta^{X,\tau_k^X}$
        & Joint flow time, vector field, and clean-endpoint predictor at solver evaluation $k$.\\
        $\mathbf Y_{t,k}^X,\mathbf W_{t,k}^X,\mathbf P^X$
        & Joint feedback target, block support/confidence operator, and block-diagonal modality preconditioner.\\
        $\mathbf J_{t,k}^X,\mathbf J_{QR}$
        & Joint clean-endpoint Jacobian and its output-modality/input-modality block, where $Q,R\in\{Z,A\}$.\\
        $\mathbf P,P_Z,P_A,k_p$
        & Block-diagonal modality preconditioner, its state and action blocks, and the proportional state-feedback gain used after modality-scale balancing.\\
        $\otimes,\mathbb 1[\cdot]$
        & Kronecker product and indicator function.\\
        \hline
    \end{tabular}
    \caption{Method-specific notation for the stage-wise and joint-generation FBFM formulations.}
    \label{tab:method-notation}
\end{table}

\clearpage

\section{Pseudoinverse Interpretation and Approximation Details}
\label{app:pseudoinverse}

\subsection{Generalized-Inverse Consistency}

Consider a noiseless linear measurement
\begin{equation}
    \mathbf{Y}_t=\mathbf{H}\mathbf{X}_t^1,
\end{equation}
where $\mathbf{H}$ has linearly independent rows. Its Moore--Penrose
pseudoinverse is
\begin{equation}
    \mathbf{H}^\dagger
    =
    \mathbf{H}^{\mathsf T}
    (\mathbf{H}\mathbf{H}^{\mathsf T})^{-1}.
\end{equation}
For a predicted clean endpoint $\hat{\mathbf{X}}_t^1$, lifting the measurement
residual back to the generation space gives
\begin{equation}
    \mathbf{H}^\dagger
    \left(
        \mathbf{Y}_t-\mathbf{H}\hat{\mathbf{X}}_t^1
    \right)
    =
    \mathbf{H}^\dagger\mathbf{Y}_t
    -
    \mathbf{H}^\dagger\mathbf{H}\hat{\mathbf{X}}_t^1.
\end{equation}
The matrix $\mathbf{P}_{\mathbf H}=\mathbf{H}^\dagger\mathbf{H}$ is the
orthogonal projector onto the row space of $\mathbf H$. Consequently,
pseudoinverse guidance compares only components that are recoverable from the
measurement.

For a nonlinear measurement operator $h$, Song et al.~\cite{song2023pseudoinverse}
require a generalized inverse satisfying
\begin{equation}
    h\circ h^\dagger\circ h=h.
\end{equation}
We additionally use the reflexive consistency relation
\begin{equation}
    h^\dagger\circ h\circ h^\dagger=h^\dagger.
\end{equation}
These identities motivate the term \emph{Moore--Penrose-consistent feedback
encoder--decoder pair}. They do not assert that arbitrary nonlinear maps possess a
global Moore--Penrose inverse. Rather, on the relevant data manifold, the encoder
and decoder preserve measurable content; under local linearization, the lifting is
modeled by the Moore--Penrose pseudoinverse of the measurement Jacobian.

\subsection{Masked Inpainting as a Projection}

Let $\mathbf M_t$ select the observed coordinates of a chunk:
\begin{equation}
    \mathbf Y_t=\mathbf M_t\mathbf X_t^1.
\end{equation}
When the rows of $\mathbf M_t$ are selected canonical basis vectors, they are
orthonormal and
\begin{equation}
    \mathbf M_t^\dagger=\mathbf M_t^{\mathsf T}.
\end{equation}
Define the full-space binary mask weights $\mathbf w_t\in\{0,1\}^D$ and their
diagonal operator $\mathbf W_t$ by
\begin{equation}
    \mathbf W_t
    =
    \operatorname{Diag}(\mathbf w_t)
    =
    \mathbf M_t^{\mathsf T}\mathbf M_t
\end{equation}
and the zero-filled observation by
$\bar{\mathbf Y}_t=\mathbf M_t^{\mathsf T}\mathbf Y_t$. Then
\begin{align}
    \mathbf M_t^\dagger
    \left(
        \mathbf Y_t-\mathbf M_t\hat{\mathbf X}_t^1
    \right)
    &=
    \mathbf M_t^{\mathsf T}\mathbf Y_t
    -
    \mathbf M_t^{\mathsf T}\mathbf M_t\hat{\mathbf X}_t^1\\
    &=
    \mathbf W_t
    \left(
        \bar{\mathbf Y}_t-\hat{\mathbf X}_t^1
    \right).
\end{align}
Hence, binary coordinate inpainting is exactly a Moore--Penrose projection.
Replacing $\mathbf w_t\in\{0,1\}^D$ with
$\mathbf w_t\in[0,1]^D$, while retaining
$\mathbf W_t=\operatorname{Diag}(\mathbf w_t)$, yields a confidence-weighted
relaxation; the resulting operator is generally no longer an orthogonal projector.

\subsection{Aligned-Coordinate Approximation}

The exact lifted discrepancy used by pseudoinverse guidance is
\begin{equation}
    \mathbf e_{\mathrm{exact}}
    =
    \mathbf W_t
    \left[
        h^\dagger(\mathbf Y_t)
        -
        h^\dagger\!\left(h(\hat{\mathbf X}_t^1)\right)
    \right].
\end{equation}
FBFM uses the aligned-coordinate approximation
\begin{equation}
    \mathbf e_{\mathrm{approx}}
    =
    \mathbf W_t
    \left[
        h^\dagger(\mathbf Y_t)
        -
        \hat{\mathbf X}_t^1
    \right].
\end{equation}
Define the masked encoder--decoder reconstruction error
\begin{equation}
    \boldsymbol\delta_t
    =
    \mathbf W_t
    \left[
        h^\dagger\!\left(h(\hat{\mathbf X}_t^1)\right)
        -
        \hat{\mathbf X}_t^1
    \right].
\end{equation}
The two discrepancies satisfy
\begingroup
\small
\begin{equation}
    \begin{aligned}
    \mathbf e_{\mathrm{exact}}
    &=
    \mathbf e_{\mathrm{approx}}-\boldsymbol\delta_t,
    &
    \left\|
        \mathbf e_{\mathrm{exact}}-\mathbf e_{\mathrm{approx}}
    \right\|_2
    &=
    \|\boldsymbol\delta_t\|_2.
    \end{aligned}
\end{equation}
\endgroup
The approximation is therefore exact whenever $h^\dagger\circ h$ is the identity
on the mask-selected coordinates, and its discrepancy is controlled directly by the
masked reconstruction error otherwise. This condition is plausible when feedback
values and model predictions have already been encoded into the same latent/action
coordinates, but it is not guaranteed for an arbitrary nonlinear encoder--decoder
pair.

\subsection{Few-Step Flow-Matching Guidance}

With
\begin{equation}
    \hat{\mathbf X}_t^1
    =
    f_\theta^\tau(\mathbf X_t^\tau)
    =
    \mathbf X_t^\tau
    +(1-\tau)v_\theta(\mathbf X_t^\tau,\tau;\mathcal H_t),
\end{equation}
the correction is computed as the VJP
\begin{equation}
    \mathbf g_t^\tau
    =
    \left(
        \frac{\partial f_\theta^\tau(\mathbf X_t^\tau)}
             {\partial\mathbf X_t^\tau}
    \right)^{\mathsf T}
    \mathbf e_t^\tau.
\end{equation}
A practical weight schedule for few-step Flow-Matching solvers is
\begin{equation}
    \lambda_\tau
    =
    \min\!\left(
        \beta,\frac{1-\tau}{\tau r_\tau^2}
    \right),
    \qquad
    r_\tau^2
    =
    \frac{(1-\tau)^2}{\tau^2+(1-\tau)^2},
\end{equation}
where $\beta$ clips the weight for numerical stability
\cite{black2025realtime}. Reverse-mode automatic differentiation evaluates the VJP
through $f_\theta^\tau$; differentiability of $h$ and $h^\dagger$ is not required.

\clearpage

\section{Implementation Details for Stage-Wise and Joint WAMs}
\label{app:implementation}
\label{app:lingbot-integration}

\subsection{Audited Boundary and Common Controls}

The corrected DreamZero integration is on branch
\texttt{experiment/dreamzero-l1mass-state-weight}. Its numerical implementation
is frozen at commit \texttt{cb08c9e}, with the associated audit and
implementation record at commit \texttt{0f2cc4f}. These revisions supersede the
earlier A6000 snapshot \texttt{a7dcd4a} for all DreamZero implementation claims.
The default \texttt{main} branch does not contain this corrected path and must
not be used to reproduce the track.

The LingBot-VA/RoboTwin2.0 implementation was audited from the current experiment
worktree based on commit \texttt{e482dcc}. Its run-specific launcher and
task-manifest changes are still under validation and have not yet been frozen
into a published commit; the final result artifacts must record that eventual
revision. Both routes use BF16 inference and retain their pretrained
transformer, VAE, text encoder, classifier-free guidance, scheduler, and cache
behavior. FBFM adds no learned module. Gradients are enabled only for the
current noisy sample while computing an endpoint VJP, and every corrected
velocity and solver sample is detached before the next step.

Each route exposes one solver path with three mask settings. The \texttt{NONE}
mode zeros the state and action masks, \texttt{RTC} retains only the
previous-action mask, and \texttt{FBFM} uses the same action mask together with
the dynamic state mask. Targets may be maintained in all modes, but a zero mask
makes them numerically inactive. This keeps the model, normalization, random
initialization, solver budget, and pseudo-asynchronous schedule fixed within
each architecture.

Table~\ref{tab:fbfm-instantiation-settings} consolidates the principal runtime
settings. Subsequent sections define their tensor alignment and scheduling
semantics in detail.

\begin{table}[H]
    \centering
    \small
    \setlength{\tabcolsep}{6pt}
    \begin{tabular}{lcc}
        \toprule
        Setting & LingBot-VA & DreamZero \\
        \midrule
        Generation & Stage-wise & Joint \\
        $(H,d,s)$ & $(32,16,16)$ & $(16,8,8)$ \\
        Predicted state slots & 2 & 2 \\
        Guided / $J$ refreshes & $25_Z/50_A$ & 16 UniPC / 8 DiT--$J$ \\
        Pseudo-clock release & 26 calls / 16 actions & 8 blocks / 8 actions \\
        State-target refresh & 4 observations / latent & Every 3 actions \\
        State preconditioner & 1 & $P_Z=56/9600$ \\
        Proportional state gain & n/a & $k_p=0.0486968$ \\
        Guidance clip $\beta$ & 10 & 10 \\
        Precision & BF16 & BF16 \\
        \bottomrule
    \end{tabular}
    \caption{Runtime settings for the two FBFM instantiations. The LingBot-VA
    pseudo-clock count includes its final cache-only video call.}
    \label{tab:fbfm-instantiation-settings}
\end{table}

\subsection{LingBot-VA on RoboTwin2.0}

\subsubsection{Tensor and Constraint Alignment}

The RoboTwin2.0 configuration predicts two video-latent slots. Each slot is paired
with 16 low-level actions, giving $H=32$. For the three-camera input, the
overhead image is resized to $256\times320$, the two wrist images to
$128\times160$, and their encoded features are spatially concatenated in the
same layout as the released model. The resulting video sample has shape
$[1,48,2,24,20]$; its state mask has shape $[1,1,2,1,1]$ and is broadcast over
each complete channel--spatial latent block.

The action solver uses a tensor of shape $[1,30,2,16,1]$. The 16 active
command channels of RoboTwin2.0 are restored to LingBot-VA's 30-channel internal layout,
normalized with the checkpoint's 1st and 99th action quantiles, and masked so
that unused internal channels remain zero. We use $d=s=16$. The last 16 actions
of the preceding chunk are placed at the first 16 temporal coordinates of the
new chunk and remain fixed throughout its generation. Since $d=H-s$, the
generalized soft-overlap interval is empty in this experiment; the realized
constraint is a hard action prefix.

State feedback is encoded with the frozen streaming VAE and compared directly
with the predicted clean endpoint in normalized latent coordinates. Hence the
implementation uses the aligned-coordinate approximation
$h^\dagger(h(\hat{\mathbf X}))\approx\hat{\mathbf X}$ discussed in
Appendix~\ref{app:pseudoinverse}, without decoding and re-encoding the predicted
endpoint at every solver step. The feedback encoder has an independent
streaming cache, so encoding a dynamic measurement does not advance the
real-history encoder.

\subsubsection{Solver and Feedback Schedule}

LingBot-VA performs stage-wise generation. The released 25-step video flow and
50-step action flow are retained. Each stage also appends a zero-time
transformer call to update its prediction cache; this call is executed under
\texttt{no\_grad} and its numerical output is not integrated. The realized
schedule is therefore 25 numerical video updates plus one cache-only
evaluation, followed by 50 numerical action updates plus one cache-only
evaluation. The guidance coefficient follows the schedule in
Section~\ref{sec:method} and is clipped at $\beta=10$.

RoboTwin2.0 execution is coupled to the video stage by a deterministic clock. The
client launches the new chunk, executes the 16-action suffix of the preceding
chunk, and releases exactly 26 video evaluations over these 16 simulator
transitions. An observation package is sampled after every four actions. The
stream is first primed with the solver-start observation; four subsequent
packages yield one normalized latent slot. Thus, under the evaluated temporal
compression, the first complete dynamic state target is formed after action
16. The first 15 transitions have released 24 evaluations, and the final
transition releases the remaining two. Feedback is consumed before the first
of these evaluations, so it affects the last numerical video update; the final
cache-only call then records the corrected state context. The action stage
starts only after this video schedule completes.

Inference and feedback use separate communication lanes. The communication
thread only queues CPU observations; queued items are encoded and installed at
video-solver boundaries. A versioned chunk context rejects duplicate or
out-of-range state slots and supplies an atomic target--mask snapshot to every
evaluation. When distributed inference is enabled, rank zero broadcasts the
queued feedback batch before all ranks update the same context version.

\subsubsection{Real-History Promotion}

The model's real KV history contains only observation--action pairs available
when the active solver was launched. Observations collected while that solver
runs first enter the dynamic FBFM feedback set. At chunk handoff, those
observations and the aligned executed action frame are staged once and written
to the real KV cache before the following launch. This one-chunk promotion rule
prevents a transition from appearing simultaneously as solver-start history
and newly arrived state feedback. The implementation also checks that the
observation and action frame counts agree before every real-history update.

\subsection{DreamZero on LIBERO}

\subsubsection{Joint Endpoint Guidance}

DreamZero predicts two video-latent slots and $H=16$ actions in one joint
solver. At a native DiT evaluation, the video sample has shape
$[1,48,2,10,20]$, while the action sample and target have shape $[1,16,32]$.
Physical LIBERO commands have seven dimensions; the preceding eight actions
are quantile-normalized, padded to 32 model coordinates, and placed in the
first eight temporal positions. The active action mask therefore contains
$8\times7=56$ scalar coordinates. During the eight-action overlap, real
observations support the first of the two future latent slots, containing
$48\times10\times20=9600$ coordinates.

Let UniPC index $j$ be served by the most recent native DiT evaluation $k$,
which supplies the unguided joint velocity $\mathbf v_k$ and clean-endpoint
Jacobian $\mathbf J_k$. In DreamZero's decreasing-noise convention, the runtime
computes
\begin{align}
    \hat{\mathbf X}_j^1
    &= \mathbf X_j^{\sigma_j}-\sigma_j\mathbf v_k, \\
    \mathbf e_j
    &= \mathbf P\mathbf W_j(\mathbf Y_j-\hat{\mathbf X}_j^1),
    \label{eq:dreamzero-current-residual}
\end{align}
followed by
\begin{align}
    \mathbf g_j &= \mathbf J_k^{\mathsf T}\mathbf e_j, \\
    \tilde{\mathbf v}_j
    &= \mathbf v_k-\lambda(\sigma_j)\mathbf g_j.
    \label{eq:dreamzero-guided-field}
\end{align}
Here $\mathbf X=[\mathbf Z,\mathbf A]$,
$\mathbf J_k=\partial\hat{\mathbf X}_k^1/
\partial\mathbf X_k^{\sigma_k}$, $\mathbf W_j$ is the binary state--action
support mask, and $\mathbf P$ is a separate block-diagonal modality
preconditioner. The residual and VJP are formed jointly with respect to the
state and action samples. This retains the cross-modal Jacobian blocks:
even with an action residual of zero, a state residual may produce a nonzero
correction in action coordinates.

At each native DiT evaluation, $\mathbf v_k$ and $\mathbf J_k$ are refreshed.
At skipped DiT indices, only these two native quantities are reused; the
endpoint, residual, VJP, guidance coefficient, and guided field are recomputed
from the current sample and $\sigma_j$. The solver cache therefore never stores
or recursively propagates $\tilde{\mathbf v}_j$. We use $\tau=1-\sigma$,
$r^2=\sigma^2/(\tau^2+\sigma^2)$, and
$\lambda(\sigma)=\min\{\sigma/(\tau r^2),\beta\}$ with $\beta=10$.

The action block uses $P_A=1$. For concision, the effective state block of
$\mathbf P$ in the equations above is $k_p P_Z$: $P_Z$ balances modality scale,
while the separate proportional gain $k_p$ tunes state-feedback loop gain. The
evaluated configuration uses $P_Z=56/9600$ and $k_p=0.0486968$. Although the
implementation applies their product by scaling the stored state-mask tensor,
$\mathbf W_j$, $P_Z$, and $k_p$ remain conceptually distinct. Schedule scalars
are evaluated in at least FP32, outputs are checked for finite values, and
guided fields are detached after the current update.
Appendix~\ref{app:dreamzero-gain-search} provides the stability motivation and
complete parameter search.

\subsubsection{Rolling Causal State Target}

DreamZero's causal VAE uses an anchor plus four temporally sampled images to
encode one future latent. The LIBERO checkpoint was trained with a three-action
video stride. The runtime retains every newly observed image in causal order,
but refreshes the hard latent target only at offsets aligned with this stride.
Missing left history is padded with the measured launch anchor; an observed
image is never copied forward into an unobserved future position. The first two
windows available during an eight-action overlap are therefore
$[0,0,0,0,3]$ after action 3 and $[0,0,0,3,6]$ after action 6. Only these two
offsets refresh the first latent slot in the evaluated wave; the second slot
remains unmasked because the overlap provides no aligned measurement for it.

Rolling feedback history is deliberately separate from the model's causal
inference history. The latter uses a one-frame warm-up and then the most recent
four chunk-level inference anchors, padding early history with the oldest
available frame. Per-action observations enter only the rolling feedback
history; a stride-aligned window updates the active FBFM target, and neither
operation advances the solver-start causal/KV history.

\subsubsection{Native Solver Schedule and Chunk Handoff}

We use $d=s=8$. DreamZero retains its 16-step UniPC scheduler and released DiT
cache mask, which evaluates the DiT and refreshes the endpoint Jacobian eight
times. A scheduler callback applies guidance to every UniPC update without
replacing the native velocities stored in \texttt{prev\_predictions}. At a
skipped DiT index, the callback reuses the latest native velocity and Jacobian
but recomputes the endpoint residual and VJP for the current update.

After each committed action is executed, the client submits its observation
and releases one native DiT block. One block may cover multiple UniPC indices;
eight action releases expose all eight DiT/Jacobian refreshes and all 16 guided
UniPC updates. The observation is retained immediately, while the hard state
target changes only when its offset reaches the three-action encoder stride.
After the eight releases, the generated suffix at positions 8--15 becomes the
execution chunk for the next wave.

The native synchronous DreamZero control is kept distinct from the matched
pseudo-asynchronous modes: it replans from the latest observation, executes the
first eight actions, and does not invoke the overlap, feedback, or pseudo-clock
interfaces. In contrast, matched \texttt{NONE}, \texttt{RTC}, and
\texttt{FBFM} runs all use the same eight-action overlap protocol and differ
only through their masks.

\subsection{Diagnostics and Numerical Safeguards}

Both routes record the chunk and solver-step identifiers, constraint version,
active mask sizes, feedback offsets, endpoint errors, state/action correction
norms, guidance weight, and GPU memory. DreamZero additionally records the
UniPC index and whether its endpoint Jacobian was refreshed or reused. Solver
outputs are rejected if they contain non-finite values. These records are used
to verify stride-aligned target activation, the native-only velocity-cache
invariant, and that a state-active joint VJP can generate a nonzero
action-coordinate correction. Because a reused Jacobian is only a local
linearization, trust-region, norm-clipping, or native-update fallback safeguards
remain to be evaluated before the final benchmark.

\clearpage

\section{Detailed Evaluation Records}
\label{app:detailed-results}

This appendix is the source-of-record layout for the main success-rate tables.
Each entry is reported as \emph{successes/trials (SR)}. The RoboTwin2.0 summary
treats CPU and GPU rendering as equivalent benchmark backends while retaining
the source counts. Because cells contain either 10 or 20 episodes and are not
seed-paired across methods, the main comparison macro-averages task-cell rates
rather than pooling their episodes.

\subsection{LingBot-VA on RoboTwin2.0}

The selected RoboTwin2.0 evaluation contains 42 tasks under both
\texttt{demo\_clean} and \texttt{demo\_randomized}. Longer tasks outside this
set are excluded from the main evaluation. The consolidated ledger supplies
all 168 cells: 42 tasks, two configurations, and two methods. Integer successes
and trials are treated as the source of truth, and the displayed rates are
recomputed from these counts. The final CPU package provides 124 cells; an
audited GPU package provides 12 FBFM Clean cells; and the remaining 32 Base
cells retain task-level counts from the shared experiment ledger. The latter
support task-level rates but not episode-identity or seed-paired analyses.

\begin{table*}[p]
    \centering
    \scriptsize
    \setlength{\tabcolsep}{3pt}
    \renewcommand{\arraystretch}{0.88}
    \resizebox{\textwidth}{!}{%
    \begin{tabular}{lcccc}
        \toprule
        Task & Clean Base & Clean FBFM & Randomized Base & Randomized FBFM \\
        \midrule
        \texttt{handover\_block} & 0/10 (0\%) & 1/10 (10\%) & 1/10 (10\%) & 0/10 (0\%) \\
        \texttt{place\_cans\_plasticbox} & 10/10 (100\%) & 10/10 (100\%) & 10/10 (100\%) & 10/10 (100\%) \\
        \texttt{stack\_blocks\_two} & 7/10 (70\%) & 10/10 (100\%) & 7/10 (70\%) & 8/10 (80\%) \\
        \texttt{open\_laptop} & 9/10 (90\%) & 9/10 (90\%) & 9/10 (90\%) & 9/10 (90\%) \\
        \texttt{place\_bread\_basket} & 9/10 (90\%) & 5/10 (50\%) & 8/10 (80\%) & 9/10 (90\%) \\
        \texttt{place\_can\_basket} & 7/10 (70\%) & 10/10 (100\%) & 7/10 (70\%) & 9/10 (90\%) \\
        \texttt{place\_object\_basket} & 9/10 (90\%) & 8/10 (80\%) & 7/10 (70\%) & 7/10 (70\%) \\
        \texttt{put\_object\_cabinet} & 7/10 (70\%) & 6/10 (60\%) & 8/10 (80\%) & 7/10 (70\%) \\
        \texttt{shake\_bottle} & 20/20 (100\%) & 10/10 (100\%) & 20/20 (100\%) & 10/10 (100\%) \\
        \texttt{shake\_bottle\_horizontally} & 20/20 (100\%) & 10/10 (100\%) & 20/20 (100\%) & 10/10 (100\%) \\
        \texttt{dump\_bin\_bigbin} & 10/10 (100\%) & 10/10 (100\%) & 10/10 (100\%) & 10/10 (100\%) \\
        \texttt{handover\_mic} & 7/10 (70\%) & 10/10 (100\%) & 8/10 (80\%) & 8/10 (80\%) \\
        \texttt{place\_dual\_shoes} & 0/10 (0\%) & 1/10 (10\%) & 1/10 (10\%) & 1/10 (10\%) \\
        \texttt{place\_bread\_skillet} & 8/10 (80\%) & 7/10 (70\%) & 6/10 (60\%) & 7/10 (70\%) \\
        \texttt{place\_burger\_fries} & 10/10 (100\%) & 18/20 (90\%) & 9/10 (90\%) & 10/10 (100\%) \\
        \texttt{place\_empty\_cup} & 20/20 (100\%) & 20/20 (100\%) & 20/20 (100\%) & 10/10 (100\%) \\
        \texttt{place\_shoe} & 5/10 (50\%) & 11/20 (55\%) & 6/10 (60\%) & 5/10 (50\%) \\
        \texttt{scan\_object} & 9/10 (90\%) & 13/20 (65\%) & 8/10 (80\%) & 8/10 (80\%) \\
        \texttt{adjust\_bottle} & 18/20 (90\%) & 20/20 (100\%) & 20/20 (100\%) & 10/10 (100\%) \\
        \texttt{beat\_block\_hammer} & 7/10 (70\%) & 15/20 (75\%) & 9/10 (90\%) & 9/10 (90\%) \\
        \texttt{click\_alarmclock} & 16/20 (80\%) & 20/20 (100\%) & 18/20 (90\%) & 10/10 (100\%) \\
        \texttt{click\_bell} & 20/20 (100\%) & 20/20 (100\%) & 20/20 (100\%) & 10/10 (100\%) \\
        \bottomrule
    \end{tabular}}
    \caption{Detailed LingBot-VA RoboTwin2.0 record, part I. Each cell reports
    successes/trials and the corresponding success rate.}
    \label{tab:lingbot-robotwin-detailed-a}
\end{table*}

\begin{table*}[p]
    \centering
    \scriptsize
    \setlength{\tabcolsep}{3pt}
    \renewcommand{\arraystretch}{0.88}
    \resizebox{\textwidth}{!}{%
    \begin{tabular}{lcccc}
        \toprule
        Task & Clean Base & Clean FBFM & Randomized Base & Randomized FBFM \\
        \midrule
        \texttt{grab\_roller} & 19/20 (95\%) & 20/20 (100\%) & 17/20 (85\%) & 10/10 (100\%) \\
        \texttt{lift\_pot} & 20/20 (100\%) & 20/20 (100\%) & 17/20 (85\%) & 10/10 (100\%) \\
        \texttt{move\_can\_pot} & 18/20 (90\%) & 19/20 (95\%) & 14/20 (70\%) & 9/10 (90\%) \\
        \texttt{move\_pillbottle\_pad} & 18/20 (90\%) & 20/20 (100\%) & 19/20 (95\%) & 9/10 (90\%) \\
        \texttt{move\_playingcard\_away} & 17/20 (85\%) & 10/10 (100\%) & 17/20 (85\%) & 10/10 (100\%) \\
        \texttt{move\_stapler\_pad} & 2/10 (20\%) & 4/10 (40\%) & 6/10 (60\%) & 6/10 (60\%) \\
        \texttt{pick\_diverse\_bottles} & 10/10 (100\%) & 10/10 (100\%) & 9/10 (90\%) & 8/10 (80\%) \\
        \texttt{pick\_dual\_bottles} & 18/20 (90\%) & 10/10 (100\%) & 14/20 (70\%) & 8/10 (80\%) \\
        \texttt{place\_a2b\_left} & 9/10 (90\%) & 9/10 (90\%) & 10/10 (100\%) & 10/10 (100\%) \\
        \texttt{place\_a2b\_right} & 9/10 (90\%) & 8/10 (80\%) & 9/10 (90\%) & 10/10 (100\%) \\
        \texttt{place\_container\_plate} & 20/20 (100\%) & 10/10 (100\%) & 17/20 (85\%) & 10/10 (100\%) \\
        \texttt{place\_fan} & 8/10 (80\%) & 9/10 (90\%) & 9/10 (90\%) & 9/10 (90\%) \\
        \texttt{place\_mouse\_pad} & 5/10 (50\%) & 5/10 (50\%) & 5/10 (50\%) & 5/10 (50\%) \\
        \texttt{place\_object\_scale} & 9/10 (90\%) & 10/10 (100\%) & 8/10 (80\%) & 7/10 (70\%) \\
        \texttt{place\_object\_stand} & 12/20 (60\%) & 7/10 (70\%) & 14/20 (70\%) & 9/10 (90\%) \\
        \texttt{place\_phone\_stand} & 9/10 (90\%) & 8/10 (80\%) & 8/10 (80\%) & 8/10 (80\%) \\
        \texttt{press\_stapler} & 20/20 (100\%) & 9/10 (90\%) & 18/20 (90\%) & 10/10 (100\%) \\
        \texttt{rotate\_qrcode} & 9/10 (90\%) & 9/10 (90\%) & 9/10 (90\%) & 9/10 (90\%) \\
        \texttt{stamp\_seal} & 10/10 (100\%) & 10/10 (100\%) & 9/10 (90\%) & 9/10 (90\%) \\
        \texttt{turn\_switch} & 12/20 (60\%) & 7/10 (70\%) & 13/20 (65\%) & 5/10 (50\%) \\
        \midrule
        \textbf{Completed task cells} & \textbf{42/42} & \textbf{42/42} & \textbf{42/42} & \textbf{42/42} \\
        \textbf{Macro SR} & \textbf{80.5\%} & \textbf{83.3\%} & \textbf{79.8\%} & \textbf{82.9\%} \\
        \bottomrule
    \end{tabular}}
    \caption{Detailed LingBot-VA RoboTwin2.0 record, part II. The final row gives
    the equal-weight task macro average used in the main text.}
    \label{tab:lingbot-robotwin-detailed-b}
\end{table*}

\subsection{DreamZero on LIBERO}

\begin{table*}[p]
    \centering
    \scriptsize
    \setlength{\tabcolsep}{8pt}
    \renewcommand{\arraystretch}{0.9}
    \resizebox{\textwidth}{!}{%
    \begin{tabular}{lccc}
        \toprule
        Suite & Task ID & Base & FBFM \\
        \midrule
        LIBERO-Spatial & 0 & 19/20 (95\%) & 17/20 (85\%) \\
        LIBERO-Spatial & 1 & 19/20 (95\%) & 18/20 (90\%) \\
        LIBERO-Spatial & 2 & 19/20 (95\%) & 19/20 (95\%) \\
        LIBERO-Spatial & 3 & 19/20 (95\%) & 19/20 (95\%) \\
        LIBERO-Spatial & 4 & 10/20 (50\%) & 8/20 (40\%) \\
        LIBERO-Spatial & 5 & 3/20 (15\%) & 2/20 (10\%) \\
        LIBERO-Spatial & 6 & 20/20 (100\%) & 20/20 (100\%) \\
        LIBERO-Spatial & 7 & 19/20 (95\%) & 19/20 (95\%) \\
        LIBERO-Spatial & 8 & 14/20 (70\%) & 17/20 (85\%) \\
        LIBERO-Spatial & 9 & 15/20 (75\%) & 15/20 (75\%) \\
        \textbf{LIBERO-Spatial} & \textbf{Subtotal} & \textbf{157/200 (78.5\%)} & \textbf{154/200 (77\%)} \\
        \midrule
        LIBERO-Object & 0 & 14/20 (70\%) & 17/20 (85\%) \\
        LIBERO-Object & 1 & 18/20 (90\%) & 17/20 (85\%) \\
        LIBERO-Object & 2 & 15/20 (75\%) & 12/20 (60\%) \\
        LIBERO-Object & 3 & 16/20 (80\%) & 18/20 (90\%) \\
        LIBERO-Object & 4 & 12/20 (60\%) & 9/20 (45\%) \\
        LIBERO-Object & 5 & 20/20 (100\%) & 20/20 (100\%) \\
        LIBERO-Object & 6 & 7/20 (35\%) & 9/20 (45\%) \\
        LIBERO-Object & 7 & 9/20 (45\%) & 7/20 (35\%) \\
        LIBERO-Object & 8 & 17/20 (85\%) & 16/20 (80\%) \\
        LIBERO-Object & 9 & 18/20 (90\%) & 19/20 (95\%) \\
        \textbf{LIBERO-Object} & \textbf{Subtotal} & \textbf{146/200 (73\%)} & \textbf{144/200 (72\%)} \\
        \bottomrule
    \end{tabular}}
    \caption{Detailed DreamZero LIBERO results under the delayed
    pseudo-asynchronous protocol, part I.}
    \label{tab:dreamzero-libero-detailed-a}
\end{table*}

\begin{table*}[p]
    \centering
    \scriptsize
    \setlength{\tabcolsep}{8pt}
    \renewcommand{\arraystretch}{0.9}
    \resizebox{\textwidth}{!}{%
    \begin{tabular}{lccc}
        \toprule
        Suite & Task ID & Base & FBFM \\
        \midrule
        LIBERO-Goal & 0 & 18/20 (90\%) & 17/20 (85\%) \\
        LIBERO-Goal & 1 & 20/20 (100\%) & 20/20 (100\%) \\
        LIBERO-Goal & 2 & 8/20 (40\%) & 18/20 (90\%) \\
        LIBERO-Goal & 3 & 9/20 (45\%) & 9/20 (45\%) \\
        LIBERO-Goal & 4 & 14/20 (70\%) & 9/20 (45\%) \\
        LIBERO-Goal & 5 & 13/20 (65\%) & 15/20 (75\%) \\
        LIBERO-Goal & 6 & 14/20 (70\%) & 14/20 (70\%) \\
        LIBERO-Goal & 7 & 20/20 (100\%) & 20/20 (100\%) \\
        LIBERO-Goal & 8 & 20/20 (100\%) & 20/20 (100\%) \\
        LIBERO-Goal & 9 & 1/20 (5\%) & 0/20 (0\%) \\
        \textbf{LIBERO-Goal} & \textbf{Subtotal} & \textbf{137/200 (68.5\%)} & \textbf{142/200 (71\%)} \\
        \midrule
        LIBERO-10 & 0 & 13/20 (65\%) & 10/20 (50\%) \\
        LIBERO-10 & 1 & 9/20 (45\%) & 12/20 (60\%) \\
        LIBERO-10 & 2 & 13/20 (65\%) & 14/20 (70\%) \\
        LIBERO-10 & 3 & 14/20 (70\%) & 12/20 (60\%) \\
        LIBERO-10 & 4 & 16/20 (80\%) & 18/20 (90\%) \\
        LIBERO-10 & 5 & 17/20 (85\%) & 19/20 (95\%) \\
        LIBERO-10 & 6 & 14/20 (70\%) & 14/20 (70\%) \\
        LIBERO-10 & 7 & 8/20 (40\%) & 13/20 (65\%) \\
        LIBERO-10 & 8 & 10/20 (50\%) & 5/20 (25\%) \\
        LIBERO-10 & 9 & 7/20 (35\%) & 9/20 (45\%) \\
        \textbf{LIBERO-10} & \textbf{Subtotal} & \textbf{121/200 (60.5\%)} & \textbf{126/200 (63\%)} \\
        \midrule
        \textbf{All suites} & \textbf{Total} & \textbf{561/800 (70.125\%)} & \textbf{566/800 (70.75\%)} \\
        \textbf{Coverage} & \textbf{All tasks} & \textbf{40/40 task cells} & \textbf{40/40 task cells} \\
        \bottomrule
    \end{tabular}}
    \caption{Detailed DreamZero LIBERO results under the delayed
    pseudo-asynchronous protocol, part II. The full record covers all ten tasks
    in each of LIBERO-Spatial, LIBERO-Object, LIBERO-Goal, and LIBERO-10. Each
    task cell reports successes/20 episodes and the corresponding success rate;
    LIBERO-90 is not included.}
    \label{tab:dreamzero-libero-detailed-b}
\end{table*}

\clearpage

\section{DreamZero State-Feedback Gain Search}
\label{app:dreamzero-gain-search}

\subsection{Parameterization and Stability Motivation}

DreamZero applies state and action constraints inside one joint solver. The
state residual used by the VJP is
\begin{equation}
    \mathbf e_j^Z
    =k_p P_Z\mathbf W_j^Z
    (\mathbf Y_j^Z-\hat{\mathbf Z}_j^1),
\end{equation}
where the binary support $\mathbf W_j^Z$, modality preconditioner $P_Z$, and
proportional gain $k_p$ have distinct roles. The support selects measured
coordinates, $P_Z$ balances the 9,600 active state coordinates against 56
active action coordinates, and $k_p$ controls the remaining state-feedback
loop gain. Neither $P_Z$ nor $k_p$ changes the action-overlap residual.

The off-diagonal blocks of the local joint endpoint Jacobian can transmit
corrections in both directions and may partially compensate dimensional scale
differences. They do not include the generally nonlinear physical transition
$P(s_{t+1}\mid s_t,a_t)$. The relevant closed loop is
\begin{equation}
    \mathbf e_t^Z
    \xrightarrow{\mathbf J_{ZA}^{\mathsf T}}
    \Delta\mathbf A_t
    \longrightarrow a_t
    \xrightarrow{P(\cdot\mid s_t,a_t)}
    s_{t+1}
    \xrightarrow{\mathcal O,E}
    z_{t+1}
    \longrightarrow \mathbf e_{t+1}^Z.
    \label{eq:closed-loop-state-feedback}
\end{equation}
The environment branch is delayed, nonlinear, and generally not the inverse of
the opposite Jacobian block. Consequently, reciprocal scaling inside one solver
linearization does not guarantee stability after actions are executed and their
consequences return as later state measurements.

\subsection{State-Preconditioner Screening}

We first compared three values of $P_Z$ while leaving the binary support mask
unchanged. The controlled sweep used the ten LIBERO-Spatial and ten
LIBERO-Object tasks, official initial-state IDs 0--4, and the same A6000,
checkpoint, solver, and pseudo-asynchronous protocol for all settings. This
gives 100 episodes per setting and 300 episodes in total.

\begin{table}[H]
    \centering
    \small
    \resizebox{\textwidth}{!}{%
    \begin{tabular}{lccc}
        \toprule
        $P_Z$ & Interpretation & Success & Max action norm \\
        \midrule
        $1$ & Unattenuated state residual & 0/100 (0\%) & $2.55\times10^7$ \\
        $\sqrt{56/9600}$ & RMS coordinate balance & 59/100 (59\%) & $9.03\times10^3$ \\
        $56/9600$ & L1 coefficient-mass balance & 73/100 (73\%) & 1.608 \\
        \bottomrule
    \end{tabular}}
    \caption{DreamZero state-preconditioner screening on LIBERO-Spatial and
    LIBERO-Object.}
    \label{tab:dreamzero-preconditioner-screen}
\end{table}

On the 100 paired initial states, L1-mass scaling gains 21 successes over RMS
and loses seven (exact two-sided McNemar $p=0.01254$). The unattenuated and RMS
settings also exhibit large action-norm tails. We therefore fix
$P_Z=56/9600$ for the subsequent gain search. This is an empirical
preconditioner selection, not a general stability proof.

\subsection{Logarithmic Screening of $k_p$}

The proportional gain was searched over $[0.001,100]$ on a logarithmic scale.
The initial points were $0.001$, $1$, and $100$; subsequent candidates were
geometric midpoints of the active interval. Each new screening point used
LIBERO-Spatial tasks 1 and 9, LIBERO-Object tasks 0 and 6, and official
initial-state IDs 0--4. The $k_p=1$ row reuses a matching historical A6000
control and was not rerun on the Pro6000. The planned point $0.0392419$ was not
executed and is not reported as a result.

\begin{table}[H]
    \centering
    \scriptsize
    \setlength{\tabcolsep}{4pt}
    \resizebox{\textwidth}{!}{%
    \begin{tabular}{crcccccl}
        \toprule
        Order & $k_p$ & Spatial 1 & Spatial 9 & Object 0 & Object 6 & Total & Source \\
        \midrule
        1 & 0.001 & 3/5 & 4/5 & 2/5 & 2/5 & 11/20 (55\%) & Pro6000 \\
        2 & 1 & 5/5 & 3/5 & 3/5 & 2/5 & 13/20 (65\%) & Historical A6000 \\
        3 & 100 & 0/5 & 0/5 & 0/5 & 0/5 & 0/20 (0\%) & Pro6000 \\
        4 & 0.0316228 & 5/5 & 4/5 & 5/5 & 2/5 & 16/20 (80\%) & Pro6000 \\
        5 & 0.177828 & 5/5 & 4/5 & 1/5 & 2/5 & 12/20 (60\%) & Pro6000 \\
        6 & 0.0749894 & 4/5 & 4/5 & 3/5 & 3/5 & 14/20 (70\%) & Pro6000 \\
        7 & 0.0486968 & 4/5 & 4/5 & 4/5 & 4/5 & 16/20 (80\%) & Pro6000 \\
        \bottomrule
    \end{tabular}}
    \caption{Per-task outcomes during logarithmic $k_p$ screening.}
    \label{tab:dreamzero-kp-screening}
\end{table}

These screening rates are used only to choose candidates for expansion. In
particular, the historical $k_p=1$ control is not used for same-machine
ranking.

\subsection{Expanded Paired Comparison}

The three selected candidates were expanded to eight eligible tasks with ten
official initial states per task. The first five trials of the four screening
tasks are exact reuse of the corresponding screening records; they were not
rerun or counted twice.

\begin{table}[H]
    \centering
    \small
    \setlength{\tabcolsep}{6pt}
    \resizebox{\textwidth}{!}{%
    \begin{tabular}{lcccc}
        \toprule
        Suite & Task & $k_p=0.0316228$ & $k_p=0.0486968$ & $k_p=0.0749894$ \\
        \midrule
        LIBERO-Spatial & 1 & 8/10 & 9/10 & 9/10 \\
        LIBERO-Spatial & 9 & 7/10 & 8/10 & 8/10 \\
        LIBERO-Object & 0 & 9/10 & 9/10 & 8/10 \\
        LIBERO-Object & 6 & 6/10 & 8/10 & 7/10 \\
        LIBERO-Goal & 0 & 8/10 & 9/10 & 9/10 \\
        LIBERO-Goal & 6 & 6/10 & 7/10 & 7/10 \\
        LIBERO-10 & 0 & 4/10 & 7/10 & 6/10 \\
        LIBERO-10 & 6 & 8/10 & 7/10 & 6/10 \\
        \midrule
        \textbf{Total} & -- & \textbf{56/80 (70\%)} & \textbf{64/80 (80\%)} & \textbf{60/80 (75\%)} \\
        \bottomrule
    \end{tabular}}
    \caption{Task-level outcomes for the three expanded $k_p$ candidates.}
    \label{tab:dreamzero-kp-expanded}
\end{table}

The middle gain is the best tested point estimate, but neither paired
comparison establishes statistical superiority at the 0.05 level. It is
therefore selected as an operational candidate rather than a universal optimum.

\subsection{Eligibility and Integrity}

The expanded analysis includes LIBERO-Spatial, LIBERO-Object, LIBERO-Goal, and
LIBERO-10. LIBERO-90 tasks 0 and 6 were executed during expansion but all 60
episodes are excluded because the RLinf checkpoint was not trained on that
suite. Each retained candidate has exactly 80 unique trial records. All 240
retained episodes contain finite executed actions, with maximum action L2 norms
of 1.719, 1.655, and 1.647 in ascending gain order; no benchmark-service or
server error was recorded.

The expanded run manifests record revision
\texttt{0f2cc4f-dirty-fc420bc56cf5}. The gain implementation and search
configuration were subsequently standardized at FBFM commit
\texttt{a051933}. The complete filtered
ledgers, source hashes, and verification report are retained under
\path{result/dmzr_weight/kp_search_pro6000/}.

\clearpage

\section{Auxiliary Video-Codec Diagnostic}
\label{app:wan-video-codec-diagnostic}

To isolate how temporal feedback coverage interacts with the video codec, we
additionally apply state-only FBFM to the auxiliary ball-meets-ball sequence.
This diagnostic is separate from the primary robot evaluation and exposes an
unresolved behavior that appears when feedback measurements cover only a
finite prefix of a temporally compressed video latent.

\begin{figure}[t]
    \centering
    \includegraphics[width=\columnwidth]{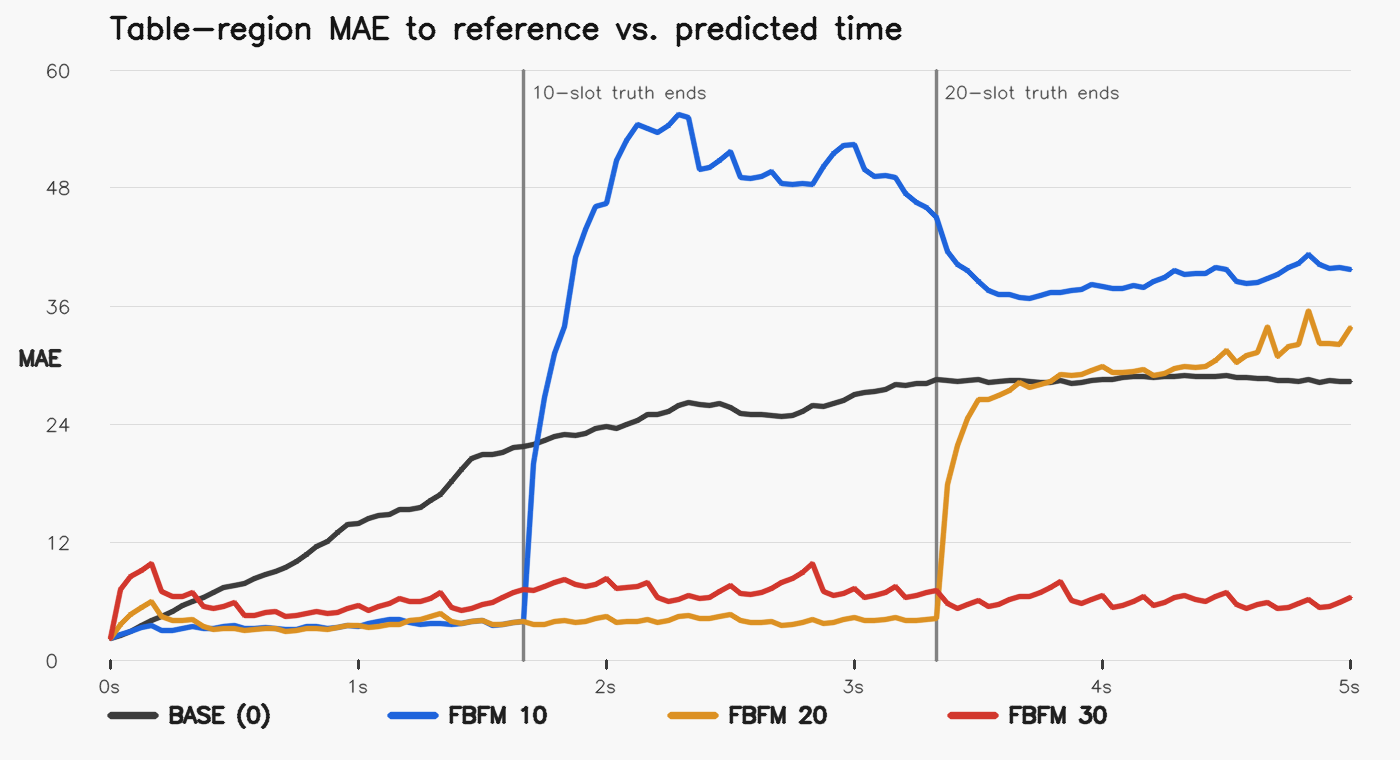}
    \caption{Image-space error under finite feedback coverage. Curves report
    frame-wise table-region MAE; vertical lines mark the ends of the measured
    prefixes for the 10- and 20-slot settings.}
    \label{fig:wan-slot-ablation-mae}
\end{figure}

Extending feedback from 10 to 30 time-aligned slots progressively covers the
measured motion. With all 30 slots, FBFM reduces table-region MAE from 21.77 to
6.34 and reduces the mean orange-ball center error from 173.84 to 3.05 pixels,
confirming that the state constraints affect the generated trajectory. The
principal failure mode is instead the large-area corruption that develops after
the finite feedback horizon in the 10- and 20-slot settings. Its onset moves
later as the measured prefix is extended, while full 30-slot coverage largely
preserves the scene structure over the evaluated horizon. Figure~\ref{fig:wan-slot-ablation-mae}
summarizes this image-space diagnostic.

The error rise aligns closely with the end of each measured prefix: the 10-slot
curve increases after approximately 1.67~s, the 20-slot curve after
approximately 3.33~s, and the 30-slot curve remains low when feedback spans the
full evaluated horizon. This curve documents a temporal association between
feedback coverage and the failure onset, but does not identify its mechanism.

One possible explanation is a Wan2.2-specific mismatch between the codec and
the generative backbone. Pseudoinverse guidance assumes that a measurement
lifted through the encoder is compatible with the model's clean latent-state
distribution. Wan2.2's joint video training and subsequent post-training may
alter the effective VAE-latent distribution used by the generator, such that
independently encoded observations do not necessarily remain semantically
meaningful clean-video states under the guided dynamics. Once the finite
constraint ends, such a mismatch could amplify and decode into large regions
without a coherent visual interpretation. We leave this as a model-specific
hypothesis: establishing the causal mechanism would require controlled
image-/latent-space diagnostics and codec retraining experiments beyond the
scope of this paper.

\begin{figure}[p]
    \centering
    \includegraphics[width=\textwidth]{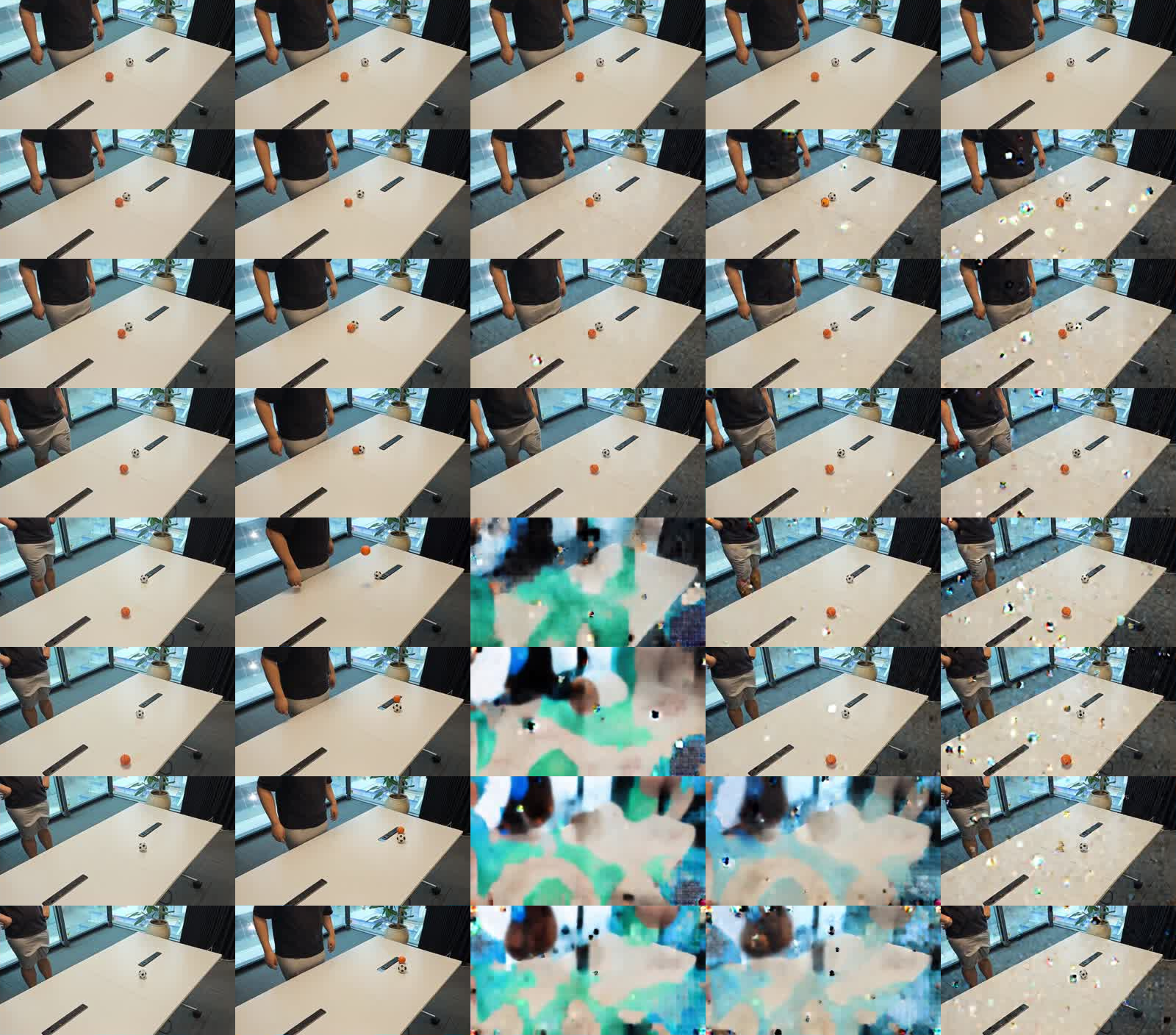}
    \caption{State-feedback coverage on the auxiliary ball-collision sequence.
    Rows show 0, 0.25, 0.5, 1, 2, 3, 4, and 5~s; columns show the reference,
    Wan2.2 Base, and FBFM with 10, 20, and 30 measured latent slots.}
    \label{fig:wan-ball-meet-ball}
\end{figure}

\end{document}